\documentclass[10pt,twocolumn,letterpaper]{article}

\usepackage{iccv}
\usepackage{times}
\usepackage{epsfig}
\usepackage{graphicx}
\usepackage{amsmath}
\usepackage{amssymb}
\usepackage{makecell}
\usepackage[linesnumbered,ruled,vlined]{algorithm2e}

\SetKwInput{KwInput}{Input}                
\SetKwInput{KwOutput}{Output}              

\SetCommentSty{mycommfont}

\usepackage{amsfonts}
\usepackage[table]{xcolor}
\usepackage{float}
\usepackage{booktabs}
\usepackage{multirow}
\usepackage{cuted}
\usepackage{capt-of}
\usepackage{xcolor}
\usepackage{bbm}

\usepackage[pagebackref=true,breaklinks=true,letterpaper=true,colorlinks,bookmarks=false]{hyperref}

\iccvfinalcopy 

\def\iccvPaperID{5754} 

\ificcvfinal\fi
 
\newif\ifdraft
\draftfalse
\drafttrue

\definecolor{orange}{rgb}{1,0.5,0}
\definecolor{magenta}{rgb}{1,0,1}
\definecolor{cyan}{rgb}{0,0.8,0.8}
\definecolor{green}{rgb}{0,0.8,0}
\definecolor{yellow}{rgb}{0.8,0.7,0.1}

\ifdraft
 \newcommand{\JB}[1]{{\color{blue}{\bf JB: #1}}}
 
 \newcommand{\NA}[1]{{\color{orange}{\bf NA: #1}}}
 
 \newcommand{\VK}[1]{{\color{magenta}{\bf VK: #1}}}

 \newcommand{\PF}[1]{{\color{red}{\bf PF: #1}}}
 
 \newcommand{\MS}[1]{{\color{green}{\bf MS: #1}}}

\else
 \newcommand{\JB}[1]{}
 
 \newcommand{\NA}[1]{}
 
 \newcommand{\VK}[1]{}
 
 \newcommand{\SC}[1]{}
 
 \newcommand{\PF}[1]{}
 
 \newcommand{\MS}[1]{}
 
\fi

\newcommand{\comment}[1]{}

\newcommand{\real}{\mathbb{R}}

\newcommand{\uvdom}{\Omega}

\newcommand{\lossmetcon}{\mathcal{L_{\text{mc}}}}

\newcommand{\losscd}{\mathcal{L}_{CD}}

\newcommand{\wmetcon}{\alpha_{\text{mc}}}

\newcommand{\ours}{OUR}

\newcommand{\atlasnet}{AN}
\newcommand{\dsr}{DSR}
\newcommand{\cyccon}{CC}
\newcommand{\nricp}{nrICP}

\newcommand{\mdist}{m_{sL2}}
\newcommand{\mrank}{m_{r}}
\newcommand{\mpckauc}{m_{\text{AUC}}}

\def\poincare/{Poincar\'e}

\newcommand{\nrm}[1]{\left\Vert#1\right\Vert}

\newcommand{\parr}[1]{\left (#1\right )}

\def\mobius/{M{\"o}bius}

\renewcommand\footnotemark{}

\begin{document}

\title{Temporally-Coherent Surface Reconstruction via Metric-Consistent Atlases \thanks{This work was partially carried out while the first author was an intern at Adobe Research and was supported in part by the Swiss National Science Foundation.}\vspace{-0.5cm}}

\author{Jan Bednarik$^1$~~~Vladimir G. Kim$^2$~~~Siddhartha Chaudhuri$^2$~~~Shaifali Parashar$^1$
\vspace{0.1cm}\\
~~~Mathieu Salzmann$^1$~~~Pascal Fua$^1$~~~Noam Aigerman$^2$
\vspace{0.2cm}\\
$^1$\small{\textsc{EPFL, Lausanne, Switzerland}~~~}$^2$\small{\textsc{Adobe Research}}
\vspace{-0.5cm}\\
}

\maketitle
\ificcvfinal\thispagestyle{empty}\fi

\begin{strip}\centering
	\vspace{-1.8cm}
	\includegraphics[width=\textwidth]{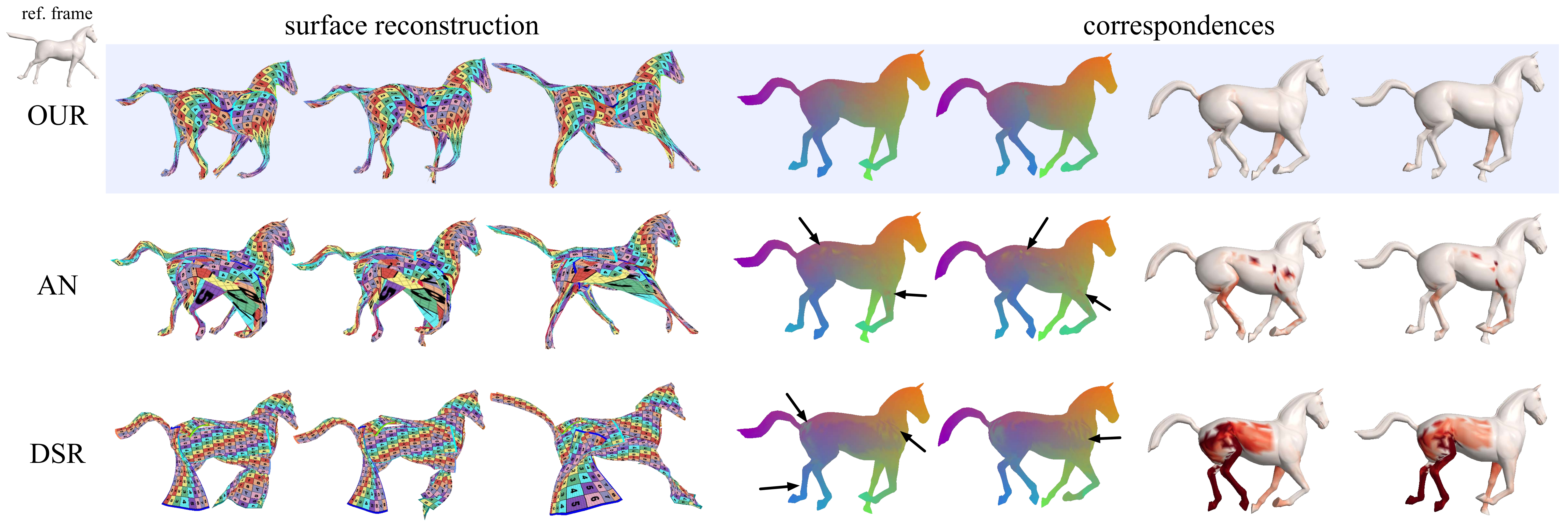}
	\captionof{figure}{{\bf Temporally-coherent reconstruction and correspondences predicted by our approach (OUR), compared with other atlas-based ones -- AtlasNet \cite{Groueix18b} (AN), and Differential Surface Representation \cite{bednarik20} (DSR).} Left: The reconstructed surfaces, textured with a consistent texture to visualize the correspondences between the surfaces. Middle and Right: deviations visualized using a colormap (middle) and a heatmap (right). The competing methods exhibit artifacts and wrong correspondences, while OUR yields reconstructions close to the GT. The black arrows point to  inconsistencies, which are absent from our results. 
	\label{fig:teaser}}
\end{strip}

\begin{abstract}
We propose a method for the unsupervised reconstruction of a temporally-coherent  sequence of surfaces from a sequence of time-evolving point clouds, yielding dense, semantically meaningful correspondences between all keyframes. We represent the reconstructed surface as an atlas, using a neural network. Using canonical correspondences defined via the atlas, we encourage the reconstruction to be as isometric as possible across frames, leading to semantically-meaningful reconstruction. Through experiments and comparisons, we empirically show that our method achieves results that exceed that state of the art in the accuracy of unsupervised correspondences and accuracy of surface reconstruction.

\vspace{-0.5cm}
\end{abstract}
\section{Introduction}
Applications such as UV-mapping, shape analysis, and partial scan-completion all rely on the availability of a surface representation that is \emph{coherent} across different instances. Namely, the different surfaces should be in correspondence, such that each point on one surface maps to a point with the same semantic meaning on another. In the literature, the most common way to achieve coherence consists of explicitly computing and establishing correspondences between non-coherent input representations, such as 3D meshes~\cite{Varanasi10,Arcila13,Roufosse19,Halimi19,Donati20,Rakotosaona20} or 3D point clouds~\cite{Insafutdinov18,Groueix19}. This, however, assumes that the input data contains points that can be matched in a semantically-meaningful manner, and in fact only circumvents the true task of retrieving a coherent surface representation.

In this paper, we tackle this problem more directly by learning to reconstruct temporally-coherent surfaces from a sequence of 3D point clouds representing a shape deforming over time. To this end, we rely on the AtlasNet patch-based representation~\cite{Groueix18b} to model the surface underlying the 3D points. However, whereas in the original AtlasNet, any patch can correspond to any part of the surface, we enforce consistency of the patch locations through the whole sequence effectively creating a time-consistent atlas. 

To learn atlases that are semantically and temporally consistent, meaning that each 2D point on each 2D atlas patch models the same semantic surface point over time, we leverage differential geometry to require the correspondences model a  close-to-isometric deformation, for which the metric tensor computed at any surface point remains constant as the shape changes. We translate this into a metric-consistency loss function, which, when minimized, implicitly establishes meaningful point correspondences.

Our approach does not require any ground-truth correspondences, which are usually difficult to obtain. Hence, it is unsupervised and can operate on any shape category {\it without} a known shape template. Yet, as shown in Fig.~\ref{fig:teaser}, it provides reliable correspondences even in cases in which the shapes are complex and the deformations are severe, unlike state-of-the-art methods which tend to break down. 

\section{Related Work}

3D temporal coherence involves both surface reconstruction and correspondence estimation, which are in interplay with one another. Both of these are well-studied, essential tasks in geometry processing, which we review next.

\paragraph{Correspondence estimation} commonly assumes that the objects are close to isometric and thus often optimizes for local distance preservation~\cite{Bronstein06,Memoli05,Salzmann09a}. This can be achieved via local shape descriptors~\cite{Ovsjanikov12,Aubry11,Sun09,Li13}, which are in turn used to obtain surface correspondences. Alternatively, obtaining correspondences can be cast as a template fitting problem~\cite{Loper15,Zuffi15}. This, however is reliant on knowing beforehand what is the class of the shape, and on having a template for this class. Simpler methods~\cite{Arcila13,Varanasi10} have been  designed for temporal registration assuming piecewise-rigidity of the shapes. However, these methods generate only region-wise correspondences. In case meshes are given, they can be parameterized into the same 2D common base-domain where correspondences can be optimized \cite{kraevoy2004cross,aigerman2014lifted,weber2014locally}. This approach relies on 3D surface (triangulations) given as input, and hence cannot be applied to point-clouds and does not reconstruct surfaces. Taking a cue from this approach, we also use a 2D domain to define the correspondences, but keep the correspondence fixed in 2D, and instead optimize the 3D surface while performing surface reconstruction.

Recently, correspondence estimation has been addressed as a learning problem. Many works use representations such as~\cite{Ovsjanikov12} to retrieve local descriptors and incorporate them in the learning process~\cite{Halimi19,Donati20,Roufosse19}. Other supervised methods have been proposed, using ground-truth correspondences as training data~\cite{Rodola14,MasBosBroVan15,Boscaini16,Monti17}.

Motivated by the fact that obtaining correspondence supervision is expensive,~\cite{Chen15,Baden18} introduced an unsupervised learning framework using  triangulated meshes. To avoid meshing,~\cite{Insafutdinov18,Groueix18a} proposed unsupervised learning techniques to extract correspondences directly from point clouds. However,~\cite{Insafutdinov18} only yields a set of semantically close points without a mechanism to find a unique correspondence, and~\cite{Groueix18a} uses a 3D template.

In contrast to existing methods, our approach yields temporally-coherent surface reconstructions from point clouds and generates meaningful point-wise correspondences. To this end, it learns a unique atlas representation similar to~\cite{Groueix18a} but enforces local metric consistency, which aims to preserve isometry at corresponding points on the output surfaces. Our method is unsupervised and does not require a shape template. Thus, the closest approach to our method is~\cite{Groueix19}, which learns correspondences by enforcing cyclic consistency across multiple shape-triplets. Our extensive comparisons with~\cite{Groueix19,Groueix18b,bednarik20} show that our method consistently outperforms these  state-of-the-art techniques.

\paragraph{Surface reconstruction} from point clouds has been thoroughly studied in geometry processing. Many non-learning techniques use mathematical tools to reconstruct the surface, e.g., solving the Poisson PDE~\cite{kazhdan2013screened}, or using Moving Least Squares~\cite{lipman2007data} to fit points to the surface; see~\cite{berger2014state} for a survey. Deep learning techniques were first successfully applied to point-cloud reconstruction~\cite{qi2016pointnet,Qi:2017:nips,Fan:2017:cvpr,Insafutdinov18}, and afterwards to surfaces, starting with the seminal AtlasNet~\cite{Groueix18a}, FoldingNet~\cite{Yang18} and their followups~\cite{Deng20,bednarik20}. Surfaces can also be reconstructed from learned elementary structures~\cite{Deprelle19}. In~\cite{williams2019deep}, an MLP was shown to be effective in reconstruction when optimized to fit a point cloud. 
 
Other representations such as meshes~\cite{Kanazawa18,pan2019deep} are simple to handle, however require a predesignated triangulation, which is not versatile enough to accommodate for arbitrary shapes with different articulations. Likewise, implicit fields such as SDF's~\cite{park2019deepsdf,mescheder2019occupancy} can represent a surface accurately, however the implicit definition does not lend itself to defining correspondences. 
 
In any case, none of  these methods target temporally-coherent surface reconstruction.
 
\paragraph{Metric preservation and shape interpolation} are closely related to our approach.
Metric preservation is widely used  when a low-distortion map between shapes is required, especially in the context of shape interpolation that has long been studied in computer graphics~\cite{aaron99,alexa2000rigid,winkler2010multi}. In recent years several data-driven methods have been proposed for this task~\cite{gao2017data,cosmo20}, but they assume to be given point  correspondences and do not infer them. Closer to our work, \cite{rakotosaona2020intrinsic} discussed how to smoothly interpolate between two point clouds, without given correspondences.  However, this work focuses solely on interpolating the point clouds without generating meaningful correspondences nor producing a continuous surface.

\section{Methodology}
\subsection{Problem statement and overview}
We assume to be given as input  a temporal sequence of 3D point clouds $P_1,...,P_K$. Our output is a corresponding sequence of reconstructed surfaces $\mathcal{S}_1,...,\mathcal{S}_K$, one for each point cloud, along with a canonical bijective mapping $\Psi_{i,j}$ between the surfaces $S_i,S_j$, defining temporally-consistent point-to-point correspondences, for any point on one of the reconstructed surfaces.

 We use an atlas-based representation with multiple patches similarly to~\cite{Groueix18b}, with an atlas ${\phi}_j$ representing each surface $\mathcal{S}_j$. This immediately defines a canonical bijective map $\Psi_{i,j}$ between any two surfaces $\mathcal{S}_i,\mathcal{S}_j$ via the shared 2D domain (see Figure \ref{fig:metric_consistency_schema}). We wish to optimize the atlases so that their surfaces satisfy two properties:
 \begin{enumerate}
     \item \textbf{Fitting.} Each surface $\mathcal{S}_k$ should model the corresponding point cloud $P_k$ as closely as possible.
    \item \textbf{Temporal coherence.} Each predefined canonical bijective map $\Psi_{i,j}$ maps semantic parts of the surface correctly between frames (nose is mapped to nose). 
\end{enumerate}

Our core observation is that we can  achieve this goal in an unsupervised manner, by making the $\Psi_{i,j}$ as isometric as possible, thus encouraging the transition from one frame in the sequence to the next to preserve local shape features, and in turn making the reconstructions consistent. We elaborate on the above next.

\subsection{Atlas-based surface representation}

\paragraph{Atlases and canonical surface correspondences.} In its most basic form, an \emph{atlas} can be defined as a map $\phi$, embedding a 2D domain $\uvdom$ to a surface in 3D $\phi:\uvdom\to \mathbb{R}^3$, such that the image of $\phi$ is $\mathcal{S}$ (we use $\uvdom = [0, 1]^2$ in all experiments). 

\begin{figure}[tb]
\begin{center}
\includegraphics[width=0.50\textwidth]{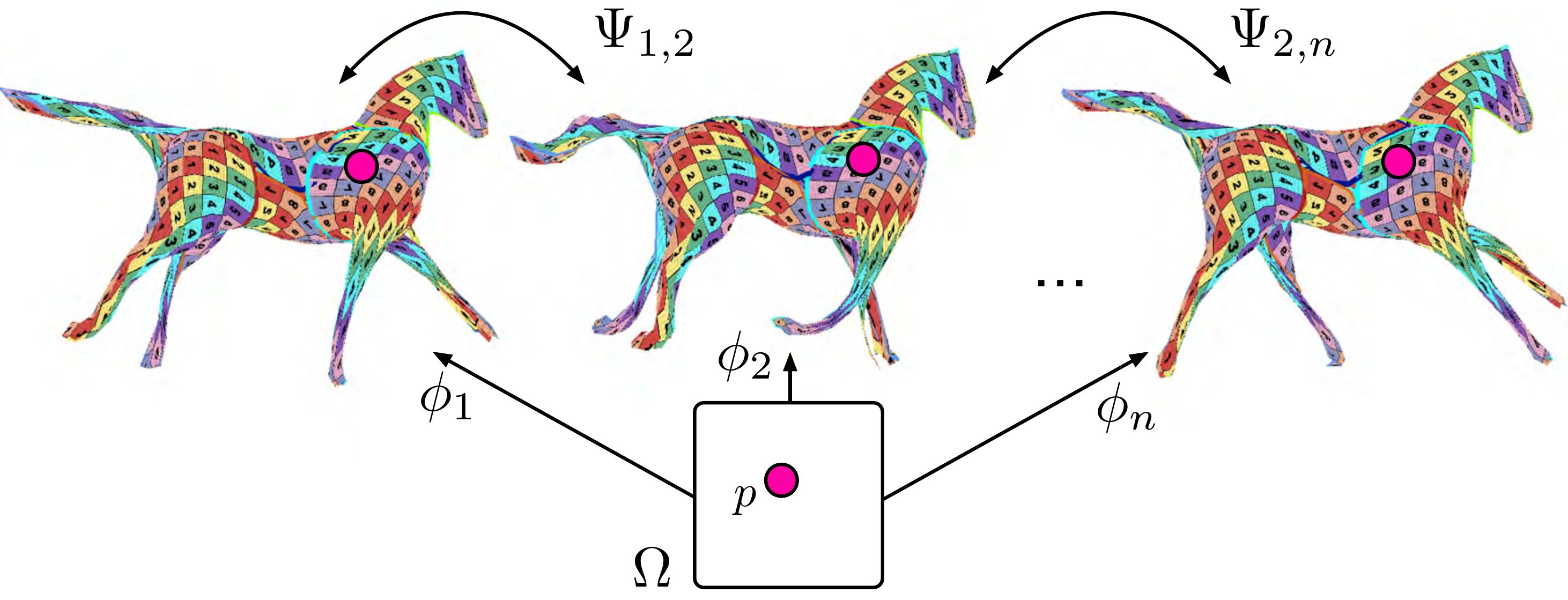}
\end{center}
\vspace{-0.4cm}
  \caption{Correspondences defined between three surfaces by the mapping of one point $p\in\Omega$ through three different atlases. }
  \vspace{-0.4cm}
\label{fig:metric_consistency_schema}
\end{figure}

Using atlases enables us to define a canonical point correspondence between any two 3D surfaces, $\mathcal{S}_1,\mathcal{S}_2$, described by two atlases, $\phi_1,\phi_2$, see Figure \ref{fig:metric_consistency_schema}. Specifically, we can trivially define a bijective (1-to-1 and onto) correspondence between the two 3D surfaces by defining the point $\phi_1\left(p\right)\in \mathcal{S}_1$ to correspond to $\phi_2\left(p\right)\in \mathcal{S}_2$, and vice versa, for any point $p\in\uvdom$. This correspondence enables us to optimize the atlases to ensure that corresponding points are mapped to the same semantic 3D surface point on $\mathcal{S}_1,\mathcal{S}_2$.

\paragraph{Isometry through metric consistency.} We enforce isometry between different atlases. To achieve that we use the \emph{Riemannian metric tensor}. For any point $p = (u, v) \in\uvdom$, the metric tensor is expressed in terms of the Jacobian, the matrix $J_\phi\in\mathbb{R}^{3\times2}$ of partial derivatives of the map $\phi$ at $p$,  $J_\phi=\left[\frac{\partial \phi}{\partial u}, \frac{\partial \phi}{\partial v}\right]$. Specifically, the metric tensor is defined as $g\left(p\right) = J_\phi\left(p\right)^{\top}{J_\phi\left(p\right)}$.   Intuitively, $g$ defines a local inner product between any two vectors  $q,r\in\mathbb{R}^2$ as $q^T\cdot g\left(p\right)\cdot r$, enabling one to measure local lengths and angles at any point $\phi\left(p\right)$ on the surface $\mathcal{S}$. 

The rest of the paper can be  understood from the high-level definition of the metric tensor as a descriptor of local geometric quantities.

Given two surfaces as above, using the canonical correspondence defined above, we can compare the metrics of the surfaces, $g_{\phi_1},g_{\phi_2}$, at corresponding points $\phi_1\parr{p},\phi_2\parr{p}$, and measure the difference between the two, $\nrm{g_{\phi_1}\parr{p}-g_{\phi_2}\parr{p}}_F$, where $\nrm{\cdot}_F$ stands for the Frobenius norm. We can now define a metric-consistency energy between the two surfaces as
\begin{equation}
    \label{eq:cons_energy}
    E_\text{cons}\parr{\phi_1,\phi_2} = \int_{p\in\uvdom} \nrm{g_{\phi_1}\parr{p}-g_{\phi_2}\parr{p}}^2_F,
\end{equation}
which measures the deviation from isometry of the map (defined by the canonical correspondence) between the two surfaces the two atlases represent. 

\subsection{Temporally-coherent surface reconstruction}
\paragraph{Atlases via a neural network.} 
To define atlases in a deep learning setting, we follow the standard AtlasNet \cite{Groueix18b}  formulation: The network receives a point $p\in\uvdom$, along with a latent code $z \in \real^{C}$ (where $C$ is the dimension of the latent space) and outputs a 3D point, essentially defining an atlas conditioned on $z$. Note that, most importantly, all differential quantities introduced in the previous section can be easily inferred for the network's atlases, since the network is a (piecewise) differentiable mapping.

Lastly, we note that instead of relying on a single map $\phi$, any  number of charts $\phi_1,\phi_2,...,\phi_M$ can be chosen before optimization, enabling mapping several 2D domains into several 3D patches, whose union forms the complete shape. This poses no change to any of the notions discussed herein, and hence we simply consider the domain $\uvdom$ and $\phi$ as aggregating all the patches, domains and maps, except when explicitly referring to these patches specifically. In all experiments, we used $M=10$ patches. 

Given a dataset with $K$ point clouds $P_1,...,P_K$, we encode each point cloud $P_k$ into a latent code $z_k$ through a PointNet \cite{Qi17a} encoder as used by \cite{Groueix18b}. We denote by $\phi_k$  the resulting atlas defined via the code $z_k$, representing the reconstructed surface.

\vspace{-1em}

\paragraph{Loss Functions.}
To enforce isometry across the sequence, we use a loss function measuring metric consistency between pairs of atlases, 
\begin{equation}
\label{eq:loss_metcon}
    \mathcal{L}_\text{metric} = \wmetcon \sum_{\left(i,j\right)\in \mathcal{I}} E_\text{cons}\parr{\phi_i,\phi_j},
\end{equation}
where $\mathcal{I}$ holds chosen pairs of surfaces out of all possible pairs, and $\wmetcon \in \real$ is a hyper-parameter of our approach.

Next, to train the network to reconstruct the given dataset, we follow standard practice in shape reconstruction \cite{Groueix18a,Deprelle19,bednarik20,Deng20} and use chamfer distance (CD) to define the \textit{reconstruction loss}
\begin{equation}
\label{eq:cd}
\begin{aligned}
    \mathcal{L_{\text{CD}}} = \frac{1}{K}\sum_{k=1}^K  &\left[\vphantom{\sum_{q\in P^k}^k}\int_{p \in \Omega}{\min_{q \in P_k}{||\phi_k(p) - q||^{2}}}\right. +  \\ & \left.\sum_{q \in P_k}{\min_{p \in \Omega}{||\phi_k(p) - q||^{2}}}\right] \;.
\end{aligned}
\end{equation}
We then take our final loss to be
\begin{equation}
    \mathcal{L} = \mathcal{L}_\text{CD} + \mathcal{L}_\text{metric}\;.
\end{equation}

\paragraph{Sampling surface pairs.} The metric consistency loss \ref{eq:loss_metcon} operates on pairs of surfaces defined by $\mathcal{I}$, $(\mathcal{S}_{i}, \mathcal{S}_{j}), (i,j)\in\mathcal{I}$. Our assumption is that the shape gradually deforms over time, and therefore surfaces in subsequent frames should change close-to-isometrically with respect to one another. Hence we define a ``time window" $\delta$, which is a hyper-parameter of our method, and sample pairs of surfaces only if they fall within that window, $(S_{i}, S_{j}): \left|i - j \right| \leq \delta $.

\subsection{Implementation details}\label{ssec:implementation_details}
Our method uses the AtlasNet~\cite{Groueix18b} architecture with the same adjustments  of~\cite{bednarik20} for computing the metric (ReLU replaced with Softplus in the decoder;  batch normalization layers  removed). We use $P = 10$ patches in all experiments. 

We use the Adam optimizer with a learning rate ${l = 0.001}$ and a batch size of $4$ for $200000$ iterations. We employ a learning rate scheduler which divides the current $l$ by a factor of $10$ at $80\%$ and $90\%$ of the training iterations. Following~\cite{Groueix18b,bednarik20}, $2500$ points are sampled from the UV domain $\uvdom$. We set the weight of the loss term $\mathcal{L}_\text{metric}$ of Eq.~\ref{eq:loss_metcon} to $\wmetcon = 0.1$, and choose the value of $\delta$ using one sequence as a validation subset and then measuring the correspondence metrics $\mdist, \mrank$ and $\mpckauc$.

At evaluation time, we follow \cite{bednarik20} and remove any patch with area smaller than $1/1000$  of the average area of a patch. We sample a given number of available points in each patch as evenly as possible using a simulated annealing based algorithm. Please refer to the supplementary material for all other details.

\section{Evaluation} \label{sec:experiments}

\begin{figure*}[th]
\begin{center}
\includegraphics[width=0.99\linewidth]{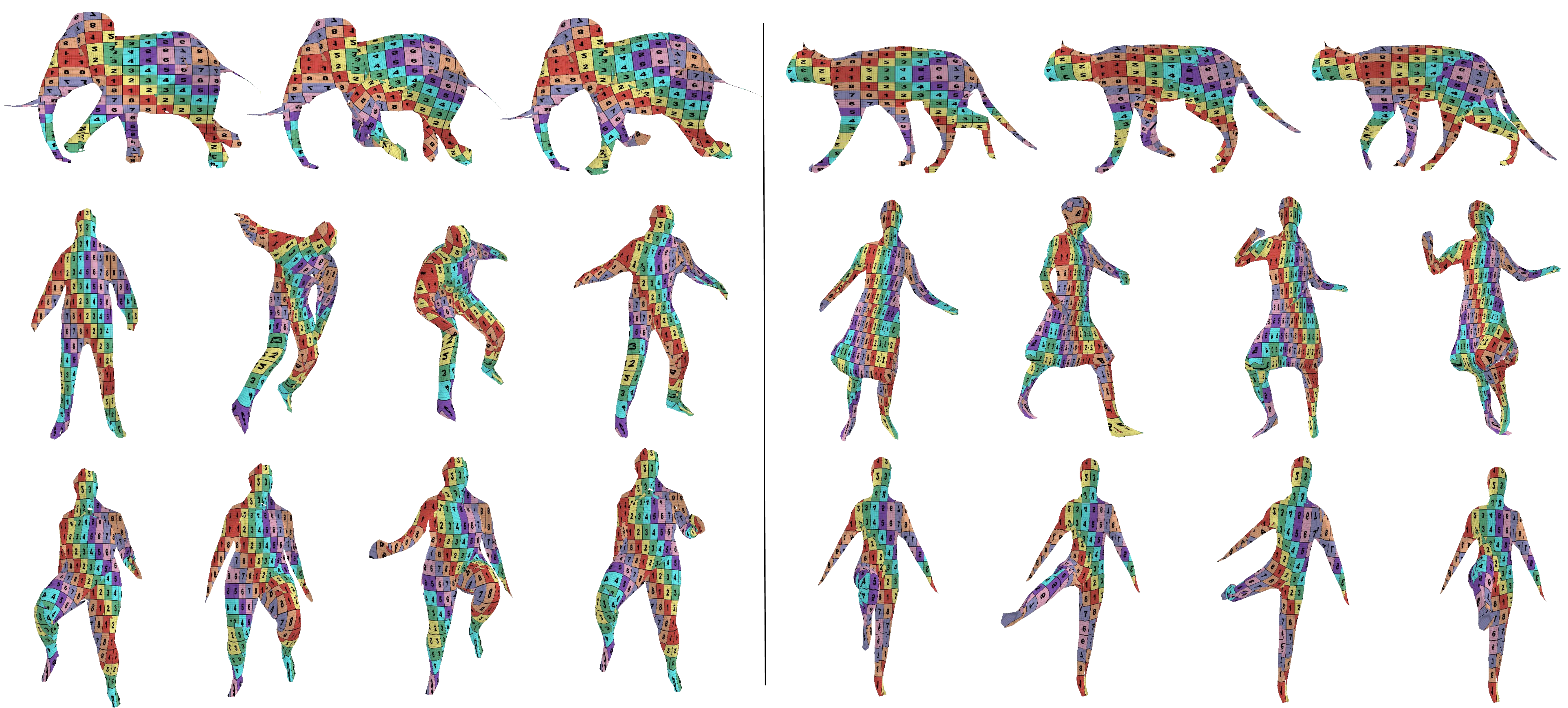}
\end{center}
  \caption{\textbf{Our temporally-coherent surface reconstructions, for 6 sequences.} (top to bottom) \texttt{elephant} and \texttt{cat} from ANIM, \texttt{jumping} and \texttt{swing} from AMA, \texttt{running\_on\_spot}  and \texttt{knees} from DFAUST. Note how the reconstructed surfaces have consistent correspondences, as well as accurate geometry.}
\label{fig:reconstructions_our}
\vspace{-0.3cm}
\end{figure*}

We tested our method by reconstructing surfaces from various raw point-cloud sequences of human and animal motions, showing our method naturally adapts to different kinds of data, without any known correspondences between the frames or a reference template shape, and without requiring prior training on any specific category.  Please refer to the supplementary material for a video showing the reconstructed sequences of all figures in the paper, as well as others, to get a full sense of the accuracy of our method. 

\paragraph{Visualization of the correspondences between surfaces.} 
Before continuing, let us explain the technique used to visualize the correspondences between the surfaces. In all figures, to illustrate the temporal consistency of our reconstructions, we use the same texture in the UV space in all frames of the sequence. Hence, corresponding regions are textured with the same checkerboard cells, revealing the accuracy of the correspondences.  

Figure~\ref{fig:reconstructions_our} shows our temporally-coherent reconstructions for six sequences. Note how our method manages to reconstruct high-curvature regions such as the elephant's tusks and the cat's tail and paws, yielding both accurate geometry and high correspondence accuracy, e.g., tracking the paws as they move. The human models exhibit much more articulated deformations, nonetheless our method tracks the limbs and maintains consistent, meaningful correspondences throughout the sequence. Please refer to the supplementary video to view the animations of the entire sequences. 

\subsection{Inferring point cloud correspondences}\label{sec:eval}
A direct application of our method is inferring point-to-point correspondences on the input point clouds. Namely, for two point clouds we map points from $P_i$ to $P_j$ via euclidean projections between the point clouds and the reconstructed surfaces, using the map $f_{i \rightarrow j} = \pi_{P_j} \circ \phi_j \circ \phi_i^{-1}  \circ  \pi_{\phi_i}$, where $\pi_\mathcal{X}$ projects a 3D point to its nearest neighbor on the surface $\mathcal{X}$ and $\phi^{-1}$ is the inverse mapping which is known implicitly. Specifically, we densely sample $N$ points in the 2D domain $\Omega$ and get their 3D counterparts via the learned $\phi$. Since $\phi$ is a bijection, we know $\phi^{-1}$ for these $N$~points.

\begin{figure}[tbh]
\begin{center}
\includegraphics[width=0.99\linewidth]{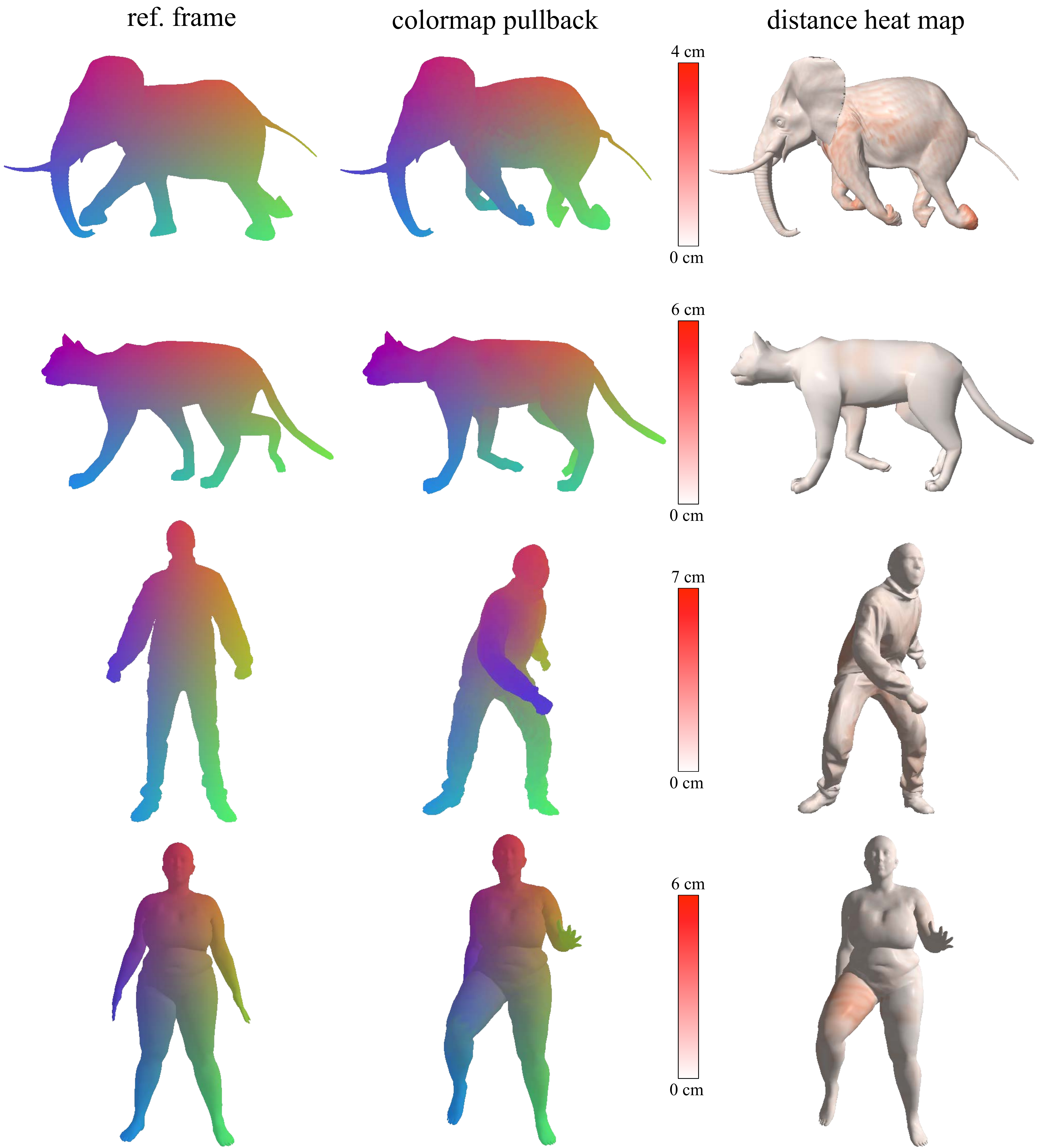}
\end{center}
  \caption{\textbf{Our correspondences retrieved on \texttt{elephant} and \texttt{cat} from ANIM, \texttt{jumping} from AMA, \texttt{running\_on\_spot} from DFAUST.} We visualize the correspondences via matching colors and show the error colorcoded as a heat map on the right. }
\label{fig:correspondence_our}
\end{figure}

We also evaluate the accuracy of our method w.r.t the ground truth correspondences of the dataset's point clouds. In Figure~\ref{fig:correspondence_our} we visualize the correspondences predicted on the input point clouds using an error colormap. We visualize of the error on the models (Note that the ground-truth triangulation of the point clouds is  only used for visualization), with red indicating the magnitude of the error -- most of the error is significantly below the maximal values chosen. Evidently, the correspondences we compute are highly accurate and exhibit small-to-no error. Some drifting can occur in relatively flat regions, such as the woman's thigh, and around very extruded regions like the elephant's feet which are harder to model exactly. 

We report quantitative evaluation of the correspondence and reconstruction in Table \ref{tab:results_all}. To evaluate the quality of correspondences, we randomly draw $M=500$ shape pairs $(P_i, P_j)$ with known ground truth correspondences $(p_k, q_k)$ where $p_k \in P_i$ and $q_k \in P_j$. Each shape has $N=3125$ points. We report the average error over $M$ pairs, with respect to the metrics described below. 

\textbf{Squared correspondence distance ($\mdist$).} This metric evaluates the error in the predicted inter-surface map $f_{i \rightarrow j}$ as $\mdist = \frac{1}{N} \sum_{k=1}^N \| f(p_k) - q_k\|^2$.

\textbf{Normalized correspondence rank ($\mrank$).} $\mrank$ expresses the rank of a predicted point with respect to all the other points on the target object. Formally $\mrank = \frac{1}{N^2}\sum_{k=1}^{N}\sum_{l=1}^{N}\mathbbm{1}_{\|q_l-q_k\|^2 < \|f(p_k)-q_k\|^2}$.

\textbf{Area under the percentage of correct keypoints (PCK) curve (\textnormal{$\mpckauc$}).} Following the literature on keypoint classification and correspondences~\cite{Huang17,You20}, we compute a mean PCK curve in a given range $[d_{\text{min}}, d_{\text{max}}]$ and report the area under that curve (AUC). We set $d_{\text{min}} = 0, d_{\text{max}} = 0.02$ in all our experiments.

\textbf{Chamfer Distance (CD).} This metric is equal to the loss term $\mathcal{L}_\text{CD}$ of Eq.~\ref{eq:cd}. Note that this is the only metric that does not evaluate the quality of correspondences but rather of the reconstruction. 

\begin{figure*}[tbh]
\begin{center}
\includegraphics[width=0.99\linewidth]{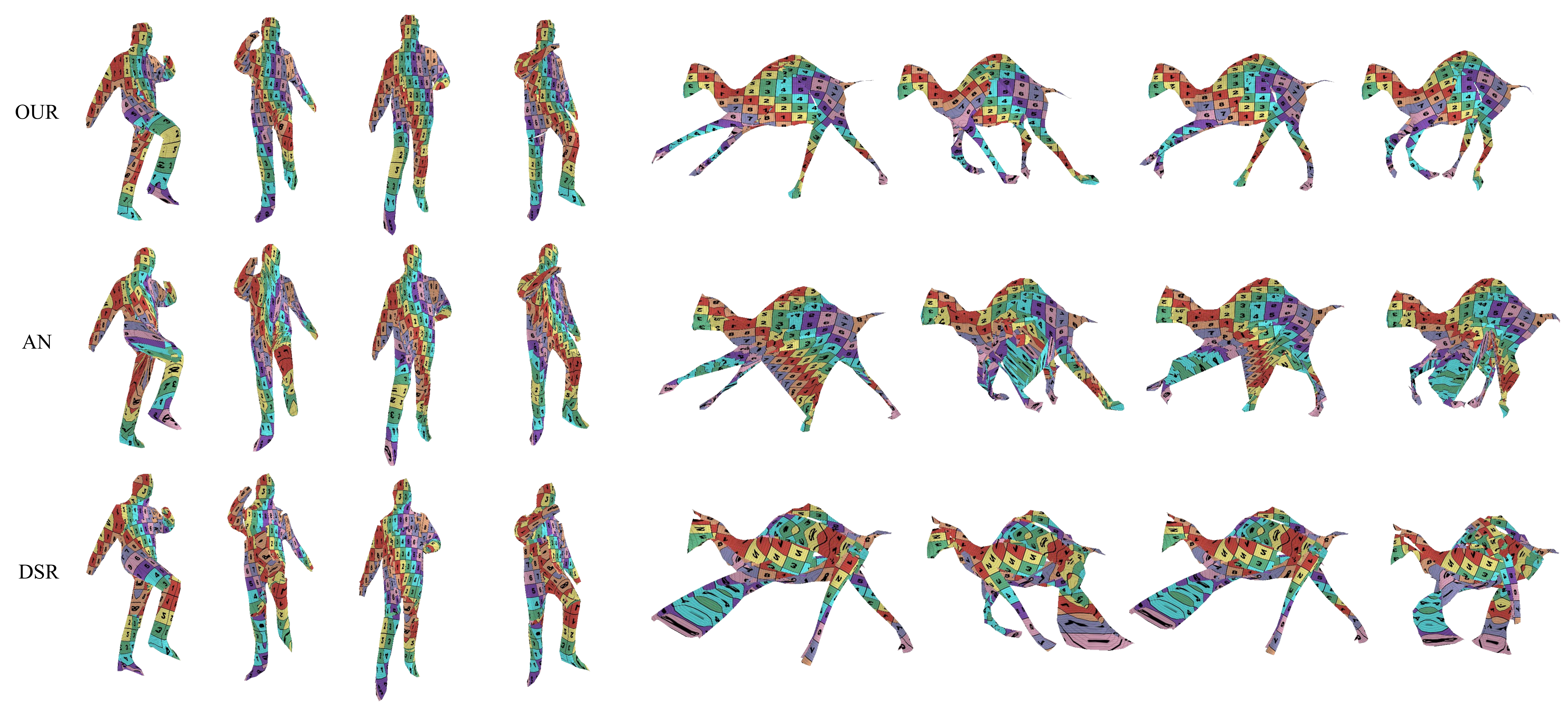}
\end{center}
  \caption{\textbf{Comparison of reconstructions of sequences \texttt{march\_1} (AMA) and \texttt{camel} (ANIM) as produced by \atlasnet{}, \dsr{} and \ours{}.} The other methods struggle to reconstruct the camel, and produce bad correspondences for the human (swapped legs on leftmost frame).}
\label{fig:comp_reconstruction}
\vspace{-0.3cm}
\end{figure*}


\subsection{Datasets}
We evaluate our method on 3 datasets of point cloud sequences.
\textbf{Animals in motion \cite{Sumner04,Aujay07} (ANIM)}   consists of 4 synthetic mesh sequences of 4-legged animals in stride. We uniformly scale each sequence s.t. the first point cloud fits in a unit cube. \textbf{Dynamic FAUST \cite{Bogo17} (DFAUST)} is a real-world dataset which contains $14$ sequences of $10$ unclothed human subjects performing various actions. \textbf{Articulated mesh animation \cite{Vlasic08} (AMA)} is a real-world dataset containing $10$ sequences depicting $3$ different human subjects performing various actions, however in contrast to DFAUST, they are wearing loose-fitting clothes making the surface more intricate and time-varying, hence more challenging for correspondence methods. We pre-processed the sequences to align them, by choosing the  rotation along the vertical axis  which minimizes the chamfer distance w.r.t. the previous frame. 

For each dataset, we use one sequence of the entire data set for validation (\texttt{cat} for ANIM, \texttt{jumping jacks} for DFAUST, \texttt{crane} for AMA). We use this validation sequence to choose the hyper-parameter $\delta$ by training our model using $\delta \in [1, 6]$ and then choosing the optimal one w.r.t the metrics $\mdist, \mrank$ and $\mpckauc$. We report the metrics only on the rest of the sequences. 

To generate point clouds from these meshes, we perform uniform random sampling  to draw $2500$ points.
We train and evaluate all methods on every sequence (e.g., walking cat or jumping human) individually.For DFAUST, we simultaneously train on all subjects performing the sequence, but still draw pairs of the same subject.

\subsection{Results and Comparisons}
\label{sec:results}
We compare our approach (\ours{}) to both traditional and deep learning based methods. 

\textbf{Non-rigid ICP} is a popular classic technique for shape registration. We use the recent implementation of ~\cite{Huang17}, which we denote as \nricp{}. We experimented with several ways to use it to match shape pairs and chose the optimal one. Please refer to the supplementary material for details.

\textbf{Atlas-based methods.} As \ours{} builds on an atlas-based representation, we compare it to the original AtlasNet \cite{Groueix18b} (\atlasnet{}). We also compare  to a more recent method \cite{bednarik20} (\dsr{}), which aims to reduce patch distortion, but is unaware of the temporal distortion.  As the base architecture of both methods is nearly identical to that of \ours{}, for fair comparison we train both methods in the same way as summarized in Section~\ref{ssec:implementation_details}.

\textbf{Cycle consistent point cloud deformation.} The recent method of~\cite{Groueix19} (\cyccon{}) learns to align  one point cloud to another in order to find correspondences. As the training of \cyccon{} relies on sampling triplets, we experimented to find the optimal sampling technique for \cyccon{} from the given sequence. Please refer to the supplementary for details.
\\~\\
All the deep learning based methods (\atlasnet{}, \dsr{}, \cyccon{}, \ours{}) are trained on the given sequence and then evaluated on it to retrieve the correspondences.

\begin{figure*}[tbh]
\begin{center}
\includegraphics[width=0.99\linewidth]{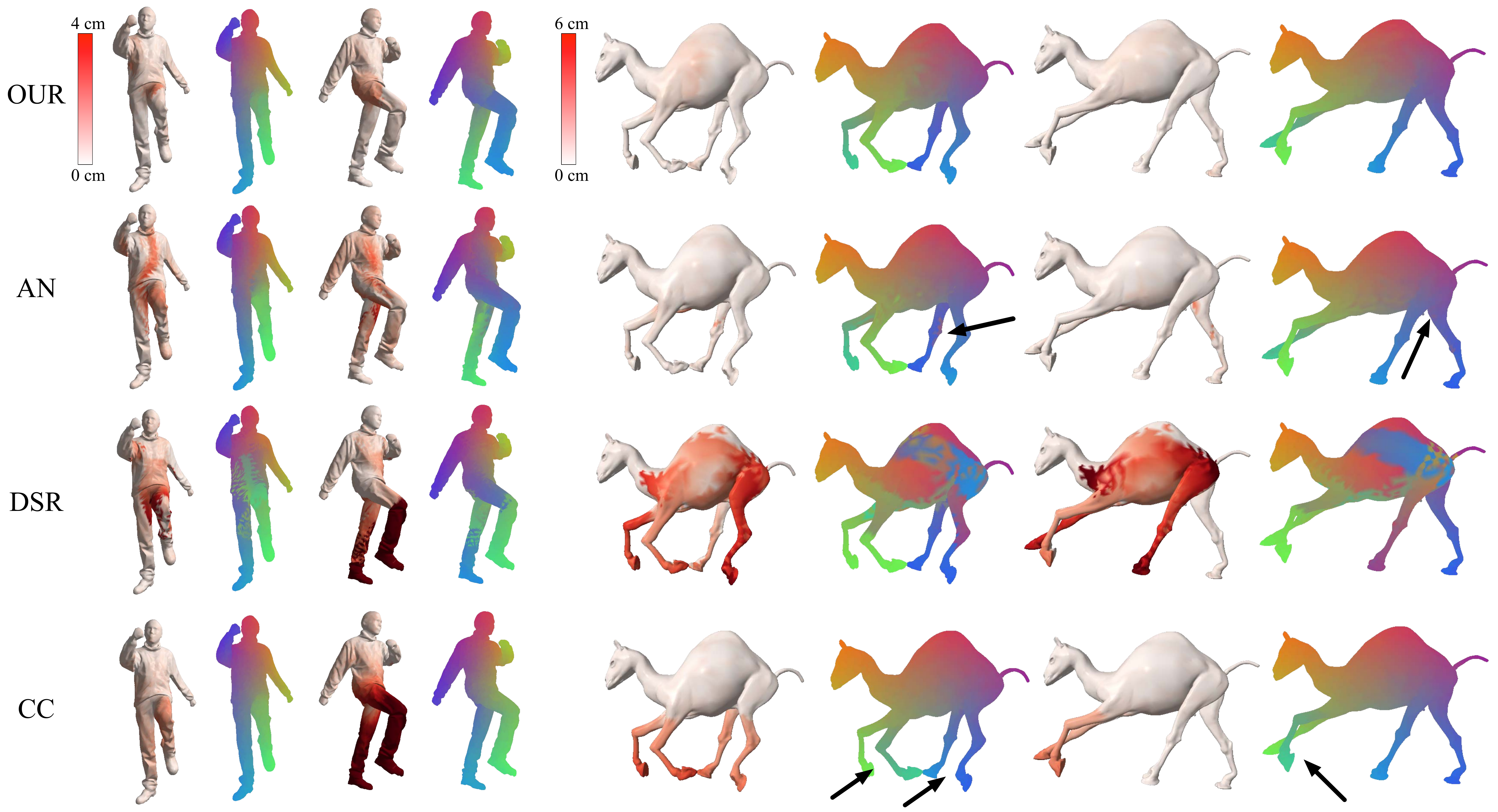}
\end{center}
  \caption{\textbf{Comparison of inferred point-cloud correspondences within 2 sequences.} The black arrows point to errors and artifacts in the correspondences, such as swapped legs of the camel as predicted by \dsr{} and \cyccon{}, and minor problems, such as small but severe local mismatches, such as the knee area of the camel as predicted by \atlasnet{}.}
  \vspace{-.2cm}
\label{fig:comp_correspondence}
\end{figure*}

Figure~\ref{fig:comp_reconstruction} shows a qualitative comparison between \ours{} and other atlas-based methods on reconstructing surfaces from point clouds, AN and DSR. As expected, in both sequences our method is more temporally-coherent and the correspondences are more accurate. Note how in the leftmost frame of the human sequence, the bending at the knee causes AN to introduce a significant amount of unnecessary distortion, while DSR maps the left leg to the right leg and vice versa. The camel sequence reveals an even more interesting observation: the temporal coherence also acts as a regularizer and makes the reconstruction itself more tight and accurate to the point cloud's geometry, as our method's reconstruction is more true to the input than the competing methods.  Please refer to the supplementary video to view the animations of the entire sequences. 

In Figure ~\ref{fig:comp_correspondence}, we show a representative qualitative comparison of the correspondences computed by \ours{} with ones inferred by the other techniques. We visualize the correspondences via matching colors, along with the measured correspondence error as a heat map.  DSR and CC swap the legs of the camel and the human. AN achieves comparable results to \ours{} on the camel, but exhibits a non-smooth jump in correspondences across the human's torso.

We report quantitative comparisons w.r.t. all metrics in Table~\ref{tab:results_all}, which demonstrates that our method achieves the best correspondence; we also achieve the  best reconstruction quality (in terms of CD) over all other methods except for AtlasNet on AMA. Since \cyccon{} and \nricp{} do not reconstruct surfaces, we do not report CD for them.
\begin{figure*}[t]
\begin{center}
\includegraphics[width=\linewidth]{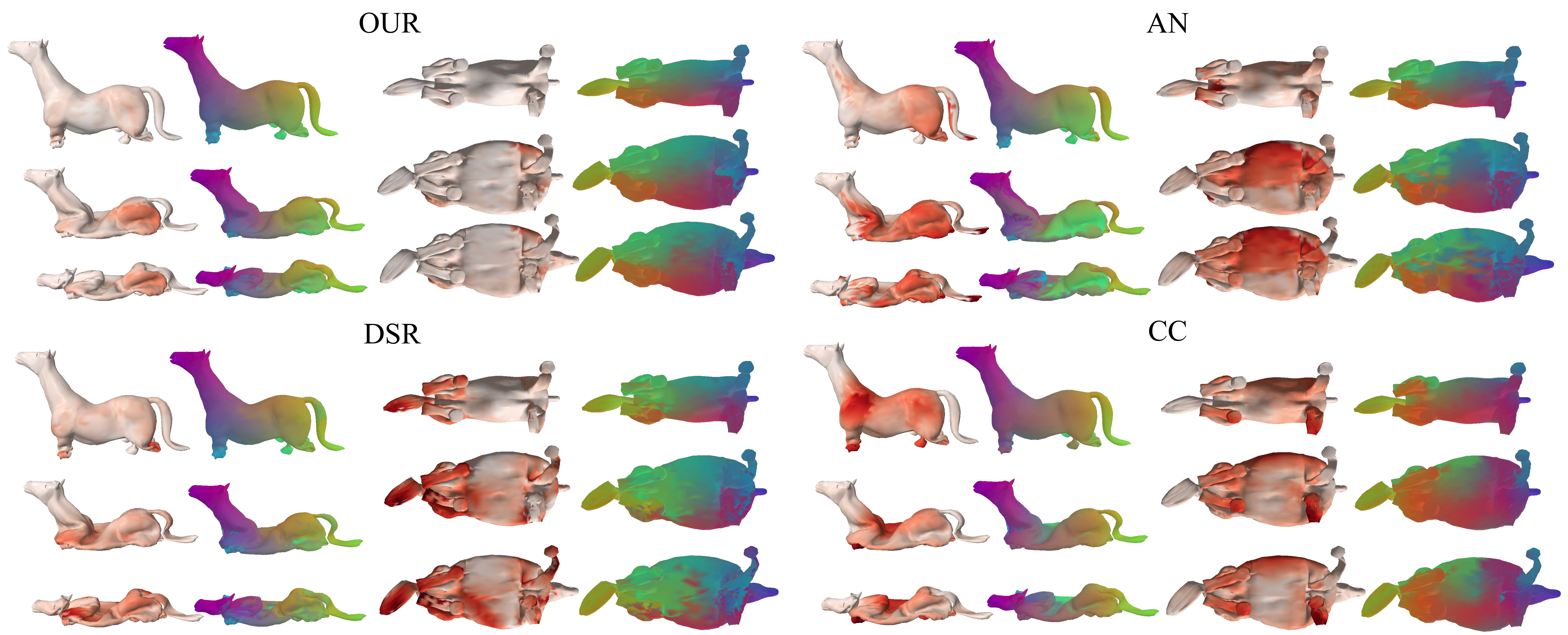}
\end{center}
  \caption{\textbf{Stress test on the sequence \texttt{horse\_collapse} from ANIM, of a horse deflating.} Despite the many self-foldovers, our method still finds accurate correspondences. The other methods fail. For each method we show  side and bottom views of 3 frames from the sequence, with correspondence visualized via matching colors, and a heatmap showing correspondence error. }
\label{fig:stresstest}
\end{figure*}
\begin{table}[tbp]
  \centering
  \caption{\textbf{Comparison of \ours{} to SotA methods on correspondence accuracy and reconstruction quality}. Our method is the most accurate and also yields reconstruction quality competitive with  \atlasnet{}.}
  \vspace{0.1cm}
    \resizebox{0.49\textwidth}{!}{
  	\begingroup
  	\setlength{\tabcolsep}{1pt}
  	\renewcommand{\arraystretch}{1.1}
    \begin{tabular}{clcccc}
    \toprule
    \textbf{dataset} & \textbf{model} & \boldmath{}\textbf{\phantom{spac}$\mdist\downarrow$\phantom{spac}}\unboldmath{} & \boldmath{}\textbf{\phantom{spa}$\mrank\downarrow$\phantom{spa}}\unboldmath{} & \boldmath{}\textbf{\phantom{sp}$\mpckauc\uparrow$\phantom{sp}}\unboldmath{} & \boldmath{}\textbf{\phantom{spac}CD $\downarrow$\phantom{spac}}\unboldmath{} \\
    \midrule
    \multirow{5}[2]{*}{ANIM} & nrICP & 70.32$\pm$84.86 & 5.46$\pm$9.52 & 74.23$\pm$13.68 & - \\
          & AN    & 18.40$\pm$24.82 & 0.78$\pm$2.85 & 96.28$\pm$1.56 & \boldmath{}\textbf{0.09$\pm$0.00}\unboldmath{} \\
          & DSR   & 46.43$\pm$67.42 & 3.44$\pm$6.71 & 83.96$\pm$9.58 & 0.19$\pm$0.01 \\
          & CC    & 33.84$\pm$54.13 & 2.21$\pm$4.75 & 87.96$\pm$7.76 & - \\
          & \textbf{OUR} & \boldmath{}\textbf{11.93$\pm$11.00}\unboldmath{} & \boldmath{}\textbf{0.30$\pm$0.57}\unboldmath{} & \boldmath{}\textbf{98.10$\pm$0.61}\unboldmath{} & \boldmath{}\textbf{0.09$\pm$0.00}\unboldmath{} \\
    \midrule
    \multirow{5}[2]{*}{AMA} & nrICP & 150.94$\pm$134.31 & 6.63$\pm$10.26 & 45.40$\pm$22.27 & - \\
          & AN    & 86.80$\pm$91.28 & 2.90$\pm$6.18 & 70.07$\pm$15.31 & \boldmath{}\textbf{0.30$\pm$0.01}\unboldmath{} \\
          & DSR   & 123.56$\pm$109.92 & 5.00$\pm$7.39 & 59.69$\pm$15.94 & 62.08$\pm$52.50 \\
          & CC    & 74.58$\pm$97.98 & 2.47$\pm$6.37 & 77.07$\pm$15.00 & - \\
          & OUR   & \boldmath{}\textbf{57.12$\pm$65.33}\unboldmath{} & \boldmath{}\textbf{1.55$\pm$3.90}\unboldmath{} & \boldmath{}\textbf{82.29$\pm$11.16}\unboldmath{} & 0.32$\pm$0.02 \\
    \midrule
    \multirow{5}[2]{*}{DFAUST} & nrICP & 79.78$\pm$118.46 & 4.09$\pm$10.17 & 74.79$\pm$15.90 & - \\
          & AN    & 31.74$\pm$43.46 & 0.90$\pm$2.95 & 91.88$\pm$5.84 & \boldmath{}\textbf{0.34$\pm$0.06}\unboldmath{} \\
          & DSR   & 68.79$\pm$61.04 & 3.76$\pm$5.19 & 78.00$\pm$6.25 & 11.21$\pm$2.89 \\
          & CC    & 29.57$\pm$65.26 & 1.12$\pm$5.26 & 94.35$\pm$9.82 & - \\
          & \textbf{OUR} & \boldmath{}\textbf{19.81$\pm$22.19}\unboldmath{} & \boldmath{}\textbf{0.38$\pm$1.17}\unboldmath{} & \boldmath{}\textbf{96.17$\pm$2.31}\unboldmath{} & \boldmath{}\textbf{0.34$\pm$0.06}\unboldmath{} \\
    \bottomrule
    \end{tabular}%
  \endgroup
	}
  \label{tab:results_all}%
  \vspace{-1em}
\end{table}%

\paragraph{Stress test.} 
In Figure~\ref{fig:stresstest}, we test the limits of our method on an extreme deformation, of a rubber horse deflating. Even under the many foldovers of the model, our method reconstructs the legs as a separate part of the surface, while the other baselines clamp different regions together, as can be seen from the bottom view.

\paragraph{Effect of the Sampling Strategy for Training Pairs.}\label{ssec:pair_sampling_strategy}
Instead of using time-adjacent point-cloud pairs, we can use random pairs instead (\textit{random}). The results of this change are shown in Table~\ref{tab:pairs_sampling_strategy}. We evaluated both strategies on the \texttt{crane}, with $\delta = 1$. The results show that \textit{neighbors} clearly yields higher correspondence accuracy. Interestingly, the deterioration in correspondence accuracy lets \textit{random} produce slightly better reconstruction in terms of CD.

\begin{table}[t]
  \centering
  \caption{\textbf{Comparison of different pair sampling strategies.} Switching from our default (\textit{neighbours}) to random leads to deterioration in correspondence accuracy and a slight improvement in CD.}
  \vspace{0.1cm}
    \resizebox{0.49\textwidth}{!}{
  	\begingroup
  	\setlength{\tabcolsep}{1pt}
  	\renewcommand{\arraystretch}{1.1}
    \begin{tabular}{ccccc}
    \toprule
    \multicolumn{1}{l}{\textbf{strategy}} & \boldmath{}\textbf{\phantom{spacer}$\mdist\downarrow$\phantom{spacer}}\unboldmath{} & \boldmath{}\textbf{\phantom{spac}$\mrank\downarrow$\phantom{spac}}\unboldmath{} & \boldmath{}\textbf{\phantom{spa}$\mpckauc\uparrow$\phantom{spa}}\unboldmath{} & \boldmath{}\textbf{\phantom{spacer}CD $\downarrow$\phantom{spacer}}\unboldmath{} \\
    \midrule
    random & 96.82$\pm$161.54 & 3.97$\pm$10.27 & 77.08$\pm$16.41 & \boldmath{}\textbf{0.30$\pm$0.01}\unboldmath{} \\
    neighbors & \boldmath{}\textbf{66.63$\pm$103.11}\unboldmath{} & \boldmath{}\textbf{2.11$\pm$6.75}\unboldmath{} & \boldmath{}\textbf{80.24$\pm$11.42}\unboldmath{} & 0.31$\pm$0.02 \\
    \bottomrule
    \end{tabular}%
    \endgroup
	}
  \label{tab:pairs_sampling_strategy}%
\end{table}%

\paragraph{Effect of the Metric Consistency Term $\mathcal{L}_\text{metric}$.}

We evaluate the effect of the hyper-parameter $\wmetcon$,  which balances metric consistency and chamfer distance. Results in terms of the correspondence metric $m_{sL2}$ are shown in Table \ref{tab:ablation_alpha_mc}, using the validation sequences of all three datasets. Setting $\wmetcon$ too low turns off $\lossmetcon$ while setting it too high overpowers $\losscd$, which imposes strict isometry and makes the position of the patches ambiguous. Hence, different values may be less or more optimal, depending on the severity of the underlying deformation. $\wmetcon \in [0.1, 1]$ yields the best results and the variations within that range are small. In all other experiments, we used $\wmetcon = 0.1$, and we note that a better, automated method to choose $\wmetcon$ may improve our performance further.

\begin{table}[t]
  \centering
  \caption{\textbf{The impact of the metric consistency term $\mathcal{L}_\text{metric}$ on the resulting accuracy of correspondences.}}
  \resizebox{0.47\textwidth}{!}{
  	\begingroup
  	\setlength{\tabcolsep}{1pt}
  	\renewcommand{\arraystretch}{0.8}
    \begin{tabular}{lcccccccc}
    \toprule
    $\wmetcon$ & \phantom{sp}$1e^{-4}$\phantom{sp} & \phantom{sp}$1e^{-3}$\phantom{sp} & \phantom{sp}$1e^{-2}$\phantom{sp} & \phantom{sp}0.1\phantom{sp} & \phantom{sp}1\phantom{sp} & \phantom{sp}$1e^{1}$\phantom{sp} & \phantom{sp}$1e^{2}$\phantom{sp} & \phantom{sp}$1e^{3}$\phantom{sp} \\
    \midrule
    \textbf{ANIM cat} & 13.2 & 11.3 & \textbf{9.8} & \textbf{9.8} & 12.5 & 14.0 & 15.1 & 62.5   \\
    \textbf{AMA crane} & 127.2 & 232.5 & 111.8 & 66.6  & \textbf{61.0}  & 102.6 & 179.3 & 174.3 \\
    \textbf{DFAUST jacks} & 35.6  & 28.0  & \textbf{23.1}  & 28.0  & 30.7  & 88.9  & 106.4 & 194.2 \\
    \bottomrule
    \end{tabular}%
    \endgroup
	}
	\label{tab:ablation_alpha_mc}%
	\vspace{-0.2cm}
\end{table}%

\section{Conclusion}
We have introduced an atlas-based method that yields temporally-coherent surface reconstructions in an unsupervised manner, by enforcing a point on the canonical shape representation to map to metrically-consistent 3D points on the reconstructed surfaces.

While our method yields better surface correspondences than state-of-the-art surface reconstruction techniques, it shares one shortcoming with these atlas-based methods. The reconstructed patches may overlap, causing imperfections in the reconstructions. Another limitation is that we use heuristics for the hyper-parameters balancing metric-consistency and reconstruction; employing an annealing-like technique which gradually permits more non-isometric deformations may be the next logical step.

We see many future applications to our approach. By replacing Chamfer distance with, e.g., some visual loss, we can apply our method to 2D sequences of images, which we believe could instigate progress in video-based 3D reconstruction. In the context of 3D geometry, our metric-consistency loss targets nearly-isometric deformations, however our framework could easily extend to other distortion measures, such as the conformal one. Studying this for non-isometric reconstruction and matching will be the focus of our future work.

{\small
\bibliographystyle{ieee_fullname}
\bibliography{egbib}
}

\newpage
\section{Supplementary Material}

\subsection{Training and Evaluation Details}
We provide details of the triplet sampling strategy used to train the cycle consistent point cloud deformation method~\cite{Groueix19} (\cyccon{}) in Section~\ref{ssec:details_of_trianing_cc}, an analysis of the strategy used to evaluate the non-rigid ICP method~\cite{Huang17} (\nricp{}) in Section~\ref{ssec:analysis_of_nricp}, more information on the points sampling strategy used to evaluate all the atlas-based methods, i.e. AtlasNet~\cite{Groueix18b} (\atlasnet{}), Differential Surface Representation~\cite{bednarik20} (\dsr{}) and our method (\ours{}), in Section~\ref{ssec:point_sampling_in_atlas_based_methods} and a time complexity analysis in Section \ref{ssec:time_complexity}.

\subsubsection{Details of Training \cyccon{}} \label{ssec:details_of_trianing_cc}

The training of \cyccon{} relies on sampling triplets of shapes from the given dataset. The authors argue that the best results were achieved when sampling triplets of shapes that are close to each other in the Chamfer distance (CD) sense. Specifically, given a randomly sampled shape $A$, two other shapes $B, C$ are randomly sampled from the $20$ nearest neighbors of $A$ to complete the triplet. Let us refer to this sampling strategy as \textit{knn}.

\ours{} itself relies on sampling shape pairs, and as shown in Section 4.3 of the main paper, better results are achieved when sampling the shape pairs from a time window $\delta$ of a given sequence (\textit{neighbors}) rather sampling pairs randomly within a sequence (\textit{random}).

For fair comparison, we experimented with training \cyccon{} using all three strategies, \textit{knn}, \textit{neighbors} and \textit{random}. Table~\ref{tab:cc_triplet_sampling} reports the results on the DFAUST dataset using the validation sequence \texttt{jumping\_jacks} and one more randomly chosen sequence \texttt{jiggle\_on\_toes}. Since \cyccon{} performs best by a large margin when trained using \textit{random}, we use this strategy for all the experiments.

\begin{table}[htbp]
  \centering
  \caption{\textbf{Comparison of different triplet sampling strategies to train \cyccon{}}. The experiments were conducted using DFAUST.}
  \vspace{0.1cm}
    \resizebox{0.49\textwidth}{!}{
  	\begingroup
  	\setlength{\tabcolsep}{1pt}
  	\renewcommand{\arraystretch}{1.1}
    \begin{tabular}{ccccc}
    \toprule
    \textbf{sequence} & \phantom{spa}\textbf{sampling}\phantom{spa} & \boldmath{}\textbf{\phantom{spacer}$\mdist\downarrow$\phantom{spacer}}\unboldmath{} & \boldmath{}\textbf{\phantom{spac}$\mrank\downarrow$\phantom{spac}}\unboldmath{} & \boldmath{}\textbf{\phantom{spa}$\mpckauc\uparrow$\phantom{spa}}\unboldmath{} \\
    \midrule
    \multirow{3}[2]{*}{jumping\_jacks} & knn   & 105.57$\pm$217.32 & 6.43$\pm$18.26 & 77.06$\pm$20.18 \\
          & neighbors & 95.13$\pm$179.81 & 6.21$\pm$17.42 & 75.84$\pm$20.78 \\
          & random & \boldmath{}\textbf{32.74$\pm$31.65}\unboldmath{} & \boldmath{}\textbf{0.70$\pm$1.68}\unboldmath{} & \boldmath{}\textbf{91.47$\pm$6.25}\unboldmath{} \\
    \midrule
    \multirow{3}[2]{*}{jiggle\_on\_toes} & knn   & 71.73$\pm$203.46 & 4.36$\pm$16.88 & 87.77$\pm$20.92 \\
          & neighbors & 47.69$\pm$99.55 & 2.15$\pm$8.48 & 88.31$\pm$12.20 \\
          & random & \boldmath{}\textbf{26.26$\pm$69.02}\unboldmath{} & \boldmath{}\textbf{0.91$\pm$5.71}\unboldmath{} & \boldmath{}\textbf{94.86$\pm$10.98}\unboldmath{} \\
    \bottomrule
    \end{tabular}%
  \endgroup
	}
  \label{tab:cc_triplet_sampling}%
\end{table}%

\subsubsection{Analysis of the \nricp{} \cite{Huang17}  Strategy} \label{ssec:analysis_of_nricp}
The \nricp{} method deforms a point cloud to best match another point cloud and thus can be used to find point-wise correspondences in an unsupervised way. Formally, let $\nu_{P_{j}}$ be the non-rigid ICP function which deforms an input point cloud $P_{i}$ to best match $P_{j}$. Following the notation introduced in Section 4.1 of the main paper, let $\pi_{\mathcal{X}}$ be a mapping that projects the points from an input point cloud 
to their respective nearest neighbors in the target point cloud $\mathcal{X}$. The simplest way to use \nricp{} to find correspondences between a pair of point clouds $(P_{i}, P_{j})$ randomly drawn from the given sequence is to compute $\pi_{P_{j}} \circ \nu_{P_{j}}(P_{i})$. Let us call this strategy \textit{random}.

Non-rigid ICP tends to break when the deformation between the two point clouds is severe. However, as we are dealing with sequences depicting a deforming shape, one can compute the correspondences between a pair of point clouds $(P_{i}, P_{j})$ by first predicting the correspondences for consecutive pairs of point clouds where the deformation is minimal, i.e., $(P_{i}, P_{i+1}), (P_{i+1}, P_{i+1}), \dots, (P_{j-1}, P_{j})$, and finally propagating the correspondences from $P_{i}$ to $P_{j}$. Formally, we compute $\pi_{P_{j}} \circ \nu_{P_{j}} \circ \dots \circ \pi_{P_{i+2}} \circ \nu_{P_{i+2}} \circ \pi_{P_{i+1}} \circ \nu_{P_{i+1}}(P_{i})$ and refer to this strategy as \textit{propagate\_simple}.

The drawback of \textit{propagate\_simple} is that every mapping $\pi_{P_{k}}$ is onto and thus throughout the propagation, progressively more source points get mapped to the same target point, which causes a loss of spatial information and ultimately yields less precise correspondences. To overcome this problem, one can replace $\pi_{P_{k}}$ with $\rho_{P_{k}}$, which performs a Hungarian matching of the input point cloud and the target point cloud $P_{k}$ with the objective of minimizing the overall per-point-pair distance. Formally, we compute $\rho_{P_{j}} \circ \nu_{P_{j}} \circ \dots \circ \rho_{P_{i+2}} \circ \nu_{P_{i+2}} \circ \rho_{P_{i+1}} \circ \nu_{P_{i+1}}(P_{i})$ and call this strategy \textit{propagate\_bijective}.

Finally, an alternative option is not to perform any projection $\pi_{P_{k}}$ or $\rho_{P_{k}}$ as we propagate the correspondences from $P_{i}$ to $P_{j}$, but instead to gradually deform the input point cloud $P_{i}$ to best match each point cloud along the sequence between $P_{i}$ and $P_{j}$. Formally, we compute $\nu_{P_{j}} \circ \dots \nu_{P_{i+2}} \circ \nu_{P_{i+1}}(P_{i})$ and refer to this strategy as \textit{propagate\_deform}.

Table \ref{tab:nricp_eval_strategy} reports the results of all four aforementioned correspondence estimation strategies on the \texttt{crane} validation sequence  from the AMA dataset. We found that \textit{propagate\_simple} suffers from the loss of spatial precision due to the onto mapping. While \textit{propagate\_bijective} overcomes this problem, the Hungarian matching introduces a strong drift along the sequence yielding even worse overall correspondences. The strategy \textit{propagate\_deform} performs the best out of all three propagation-based strategies, but is still outperformed by the simplest strategy \textit{random}. Therefore, as \textit{random} yields the highest correspondence accuracy, we use it to evaluate \nricp{} on all datasets.

\begin{table}[htbp]
  \centering
  \caption{\textbf{Comparison of strategies used to establish correspondences with \nricp{}}. The experiments were conducted on the \texttt{crane} validation sequence from the AMA dataset.}
  \vspace{0.1cm}
    \resizebox{0.49\textwidth}{!}{
  	\begingroup
  	\setlength{\tabcolsep}{1pt}
  	\renewcommand{\arraystretch}{1.1}
    \begin{tabular}{cccc}
    \toprule
    \textbf{strategy} & \boldmath{}\textbf{\phantom{spacer}$\mdist\downarrow$\phantom{spacer}}\unboldmath{} & \boldmath{}\textbf{\phantom{spac}$\mrank\downarrow$\phantom{spac}}\unboldmath{} & \boldmath{}\textbf{\phantom{spa}$\mpckauc\uparrow$\phantom{spa}}\unboldmath{} \\
    \midrule
    random & \boldmath{}\textbf{172.55$\pm$167.76}\unboldmath{} & \boldmath{}\textbf{7.83$\pm$12.56}\unboldmath{} & \boldmath{}\textbf{41.61$\pm$19.29}\unboldmath{} \\
    propagate\_simple & 211.11$\pm$147.43 & 9.38$\pm$10.32 & 23.01$\pm$18.31 \\
    propagate\_bijective & 213.87$\pm$169.00 & 10.55$\pm$13.99 & 25.31$\pm$17.80 \\
    propagate\_deform & 206.64$\pm$150.45 & 10.40$\pm$13.35 & 25.41$\pm$16.55 \\
    \end{tabular}%
  \endgroup
	}
  \label{tab:nricp_eval_strategy}%
\end{table}%

\subsubsection{Point Sampling in Atlas Based Methods} \label{ssec:point_sampling_in_atlas_based_methods}

The original AtlasNet work \cite{Groueix18b} argues that better reconstruction accuracy is achieved if the 2D points sampled from the UV domain $\uvdom$ are spaced on a regular grid. As explained in Section 4.1 of the main paper, at evaluation time each atlas based method, i.e., \atlasnet{}, \dsr{} and \ours{}, predicts $N = 3125$ points. Due to the unknown number of collapsed patches, which are discarded at runtime, it might not be possible to evenly split $N$ points into $P$ non-collapsed patches so that the points would form a regular grid in the UV space $\uvdom$.

\begin{figure}[tbh]
\begin{center}
\includegraphics[width=0.9\linewidth]{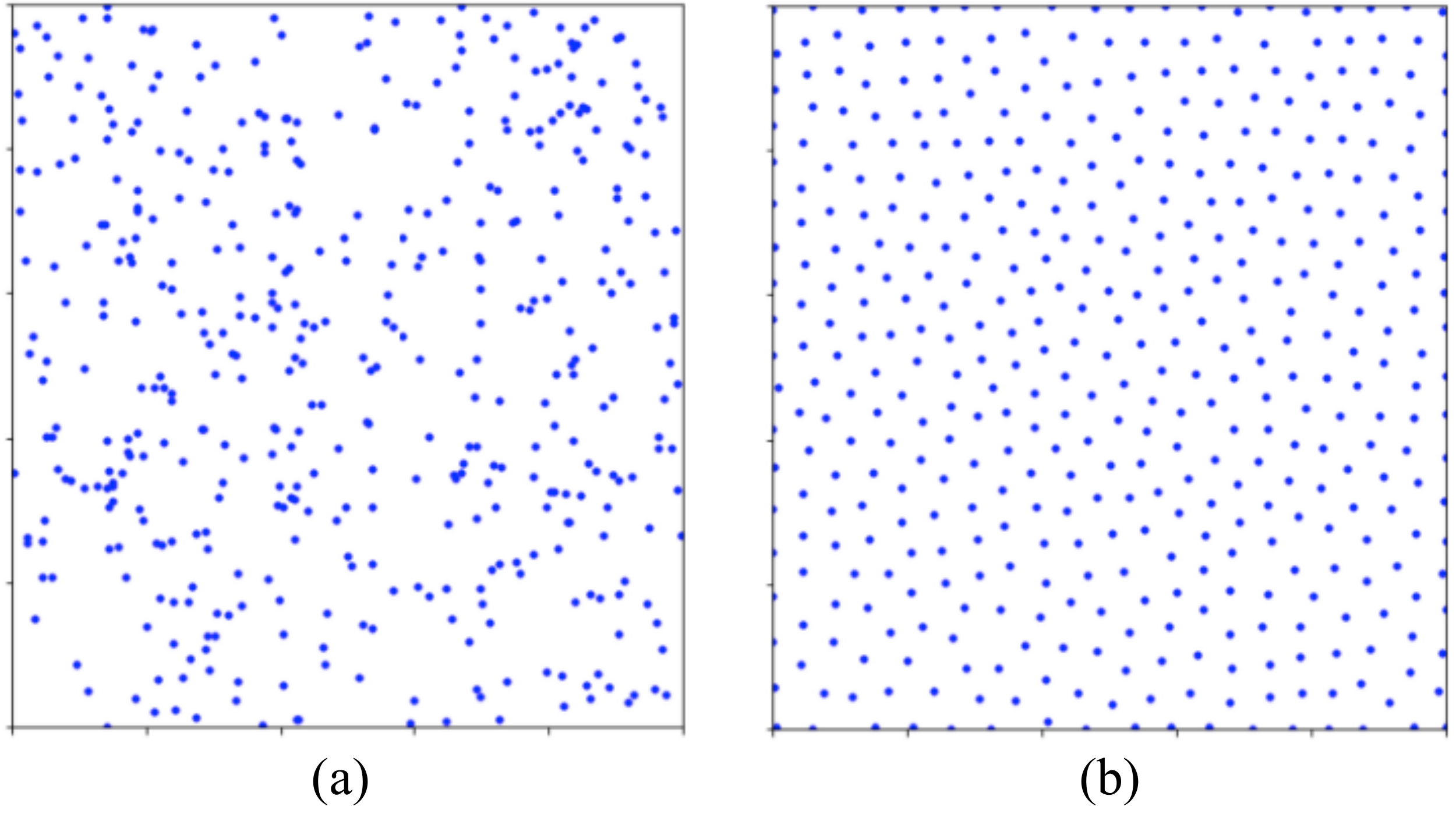}
\end{center}
  \caption{\textbf{Comparison of (a) uniform and (b) as regular as possible 2D points sampling.}}
\label{fig:regular_2d_pts}
\end{figure}

Therefore, instead of using a regular grid, we distribute the given available number of points as regularly as possible in the 2D domain using a simulated annealing based algorithm. The points are initially distributed uniformly at random, and then their position is iteratively adjusted so that every point maximizes its distance to the nearest points. This procedure is summarized in Algorithm \ref{alg:regular_2d_pts}. The difference between random and as regular as possible 2D points sampling is demonstrated in Fig. \ref{fig:regular_2d_pts}.

\begin{algorithm}[h]
\label{alg:regular_2d_pts}
\DontPrintSemicolon
  
  \KwInput{$M \in \mathbb{N}$  \tcp*{Number of 2D points.}}
  \KwOutput{$p_{i} \in \real^{2}, \forall 1\leq i \leq M$ \tcp*{2D points.}}
  \tcc{Initialization}
  step := $\frac{1}{4\sqrt{M}}$\;
  decay := 0.994\;
  $p_{i} \sim \mathcal{U}(\mathbf{0}, \mathbf{1}), \forall 1 \leq i \leq M$\tcp*{Random 2D points.}
  iter := 0\;
  \tcc{Main algorithm.}
  \While{$\text{iter} < 250$}
  {
        \For{$i:=1$ \KwTo $M$}{
      		$d_{i} :=\ \min_{j \neq i}{||p_{i} - p_{j}||}$\;
      		$\alpha_{i} \sim \mathcal{U}(0, 2\pi)$\;
      		$p_{i}^{\text{new}}$ := $p_{i} + \text{step} \cdot R(\alpha_{i})\begin{bmatrix}1 \\ 0\end{bmatrix}$\tcp*{R: rot. matrix} \; 
      		$d_{i}^{\text{new}} := \min_{j \neq i}{||p_{i}^{\text{new}} - p_{j}^{\text{new}}||}$\;
      		\If{$d_{i}^{\text{new}} > d_{i}$}
      		{
      		    $p_{i}$ := $p_{i}^{\text{new}}$
      		}
      	}
      	step := step $\cdot$ decay\;
      	iter := iter $+ 1$\;
  }
\caption{As regular as possible 2D points.}
\end{algorithm}

\subsubsection{Time Complexity} \label{ssec:time_complexity}
The optimization of all the learning based methods was performed using an Nvidia Tesla V100 GPU, and processing a sequence of average length takes $16.1$ hours for OUR, while AN, DSR and CC take $4.1$, $16.4$ and $9.7$ hours, respectively. nrICP does not involve the optimization stage and can process $\sim1$ sample per second.

\subsection{Complete Results}\label{sec:complete_results}

We provide details of the search for the best value of the hyper-parameter $\delta$ in Section \ref{ssec:tuning_time_window} and we list the complete per-sequence results of all the evaluated methods on all the datasets in Section \ref{ssec:evaluation_on_all}.  Furthermore, we refer the reader to the \href{https://youtu.be/jfNQPTsbM3g}{supplementary video}\footnote{https://youtu.be/jfNQPTsbM3g} which contains the comparison of all methods on multiple sequences from all the datasets.
    
\subsubsection{Tuning the Time-Window $\delta$}\label{ssec:tuning_time_window}
As described in Section 3.3 of the main paper, \ours{} relies on sampling pairs of shapes from a time window denoted as $\delta$. We tuned this hyper-paramater individually for every dataset using a respective validation sequence, and set it to the values yielding the best correspondence accuracy as measured by the metrics $\mdist, \mrank$ and $\mpckauc$. Table \ref{tab:search_delta} lists the results of training \ours{} for $\delta \in [1, 6]$ and justifies the selection of $\delta = 1$ for ANIM, $\delta = 1$ for AMA and $\delta = 5$ for DFAUST.

Note that, as the ANIM and AMA  datasets appear to have lower frame-rates than the DFAUST dataset, i.e., the surface undergoes larger motion from frame to frame, the correspondence error clearly decreases with the decreasing size of the time window $\delta$, indicating that our method benefits from observing pairs of shapes which are similar enough to each other. On the other hand, as the DFAUST dataset in general exhibits small frame to frame changes, the search reveals that our method can benefit from observing pairs from larger time windows, since in this case the consecutive frames are nearly identical and decreasing $\delta$ makes $\lossmetcon$ less useful. Note, however, that using the value $\delta = 1$ for all the sequences shown in this paper still consistently outperforms all the competing methods.

\begin{table}[htbp]
  \centering
  \caption{\textbf{Search for the best value of the hyper-parameter $\delta$ used by \ours{} on each dataset}.}
  \vspace{0.1cm}
    \resizebox{0.49\textwidth}{!}{
  	\begingroup
  	\setlength{\tabcolsep}{1pt}
  	\renewcommand{\arraystretch}{1.1}
    \begin{tabular}{cccccc}
    \toprule
    \textbf{dataset} & \textbf{neigh.} & \boldmath{}\textbf{\phantom{spac}$\mdist\downarrow$\phantom{spac}}\unboldmath{} & \boldmath{}\textbf{\phantom{spa}$\mrank\downarrow$\phantom{spa}}\unboldmath{} & \boldmath{}\textbf{\phantom{sp}$\mpckauc\uparrow$\phantom{sp}}\unboldmath{} & \boldmath{}\textbf{\phantom{spac}CD $\downarrow$\phantom{spac}}\unboldmath{} \\
    \midrule
    \multirow{6}[2]{*}{\makecell{ANIM\\(cat)}} & \textbf{1}     & \textbf{9.80$\pm$14.36} & 0.24$\pm$0.60 & \textbf{98.27$\pm$0.82} & 0.39$\pm$0.00 \\
          & 2     & 10.27$\pm$15.03 & 0.24$\pm$0.54 & 98.09$\pm$1.01 & 0.39$\pm$0.00 \\
          & 3     & 10.07$\pm$15.52 & \textbf{0.23$\pm$0.56} & 98.06$\pm$0.94 & \textbf{0.38$\pm$0.00} \\
          & 4     & 17.10$\pm$37.51 & 0.78$\pm$3.27 & 94.49$\pm$4.60 & 0.38$\pm$0.01 \\
          & 5     & 44.58$\pm$88.60 & 3.45$\pm$10.04 & 85.76$\pm$11.61 & 0.41$\pm$0.00 \\
          & 6     & 11.45$\pm$16.33 & 0.30$\pm$0.66 & 97.78$\pm$1.03 & 0.39$\pm$0.00 \\
    \midrule
    \multirow{6}[2]{*}{\makecell{AMA\\(crane)}} & \textbf{1}     & \textbf{66.63$\pm$103.11} & \textbf{2.11$\pm$6.75} & \textbf{80.24$\pm$11.42} & \textbf{0.31$\pm$0.02} \\
          & 2     & 99.86$\pm$163.94 & 4.31$\pm$10.82 & 76.91$\pm$17.36 & \textbf{0.31$\pm$0.02} \\
          & 3     & 91.09$\pm$138.11 & 3.68$\pm$9.22 & 74.42$\pm$16.49 & 0.32$\pm$0.01 \\
          & 4     & 81.15$\pm$130.99 & 3.02$\pm$8.67 & 77.58$\pm$13.50 & 0.33$\pm$0.01 \\
          & 5     & 106.34$\pm$166.29 & 4.61$\pm$10.85 & 74.10$\pm$17.69 & 0.34$\pm$0.02 \\
          & 6     & 113.02$\pm$162.47 & 5.16$\pm$11.91 & 68.48$\pm$20.39 & 0.35$\pm$0.09 \\
    \midrule
    \multirow{6}[2]{*}{\makecell{DFAUST\\(jumping\_jacks)}} & 1     & 32.71$\pm$46.68 & 0.92$\pm$3.15 & 91.77$\pm$4.53 & 0.51$\pm$0.09 \\
          & 2     & 32.01$\pm$51.48 & 0.89$\pm$3.50 & 92.60$\pm$3.87 & 0.48$\pm$0.11 \\
          & 3     & 29.39$\pm$33.80 & 0.73$\pm$2.25 & 93.30$\pm$2.86 & 0.50$\pm$0.09 \\
          & 4     & 30.67$\pm$45.30 & 0.92$\pm$3.32 & 92.38$\pm$3.77 & 0.55$\pm$0.15 \\
          & \textbf{5}     & \textbf{27.98$\pm$38.15} & \textbf{0.67$\pm$2.55} & \textbf{93.65$\pm$3.15} & \textbf{0.41$\pm$0.08} \\
          & 6     & 29.80$\pm$51.77 & 0.84$\pm$3.56 & 93.06$\pm$3.76 & 0.48$\pm$0.09 \\
    \bottomrule
    \end{tabular}%
  \endgroup
	}
  \label{tab:search_delta}%
\end{table}%

\subsubsection{Impact of $\wmetcon$ on the Visual Quality}\label{ssec:impact_of_alphamc_on_visual_quality}
\begin{figure}[h]
    \begin{center}
    \includegraphics[width=0.99\linewidth]{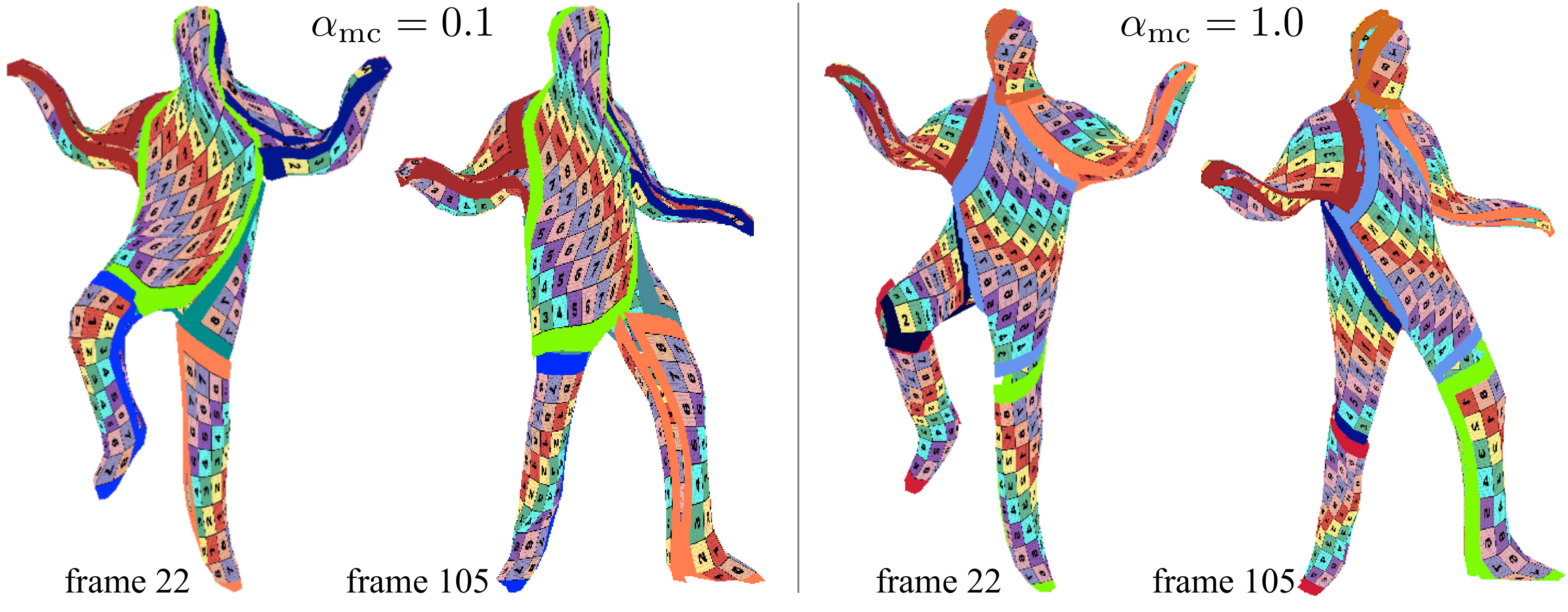}
    \end{center}
    \caption{\textbf{Comparison of the reconstruction and correspondence quality when using $\wmetcon = 0.1$ and $\wmetcon = 1.0$.} The sample pair comes from the sequence \texttt{crane} of AMA. Note that these are two independent runs, therefore, the spatial distribution of the patches is arbitrary.}
    \label{fig:alpha_mc_comparison_on_crane}
\end{figure}

As shown in Table~3 of the main paper, every dataset benefits from a different value of the hyper-parameter $\wmetcon$ which balances metric consistency and Chamfer distance, while $\wmetcon \in [0.1, 1.0]$ yields the best quantitative results. Here we show that this fact manifests in the qualitative results as well. The sequence \texttt{crane} from AMA is one case where setting $\wmetcon = 1$ instead of $0.1$ yields better quantitative results. However, both reconstructions are visually comparable, as shown in Fig. \ref{fig:alpha_mc_comparison_on_crane}.

\subsubsection{Evaluation on all Datasets and Stress Test} \label{ssec:evaluation_on_all}
For brevity, Section 4.3 of the main paper only reports the mean results computed over all the sequences contained in the individual datasets. Here we report detailed results for each sequence separately. The results of all methods evaluated on the ANIM, AMA and DFAUST datasets are summarized in Tables \ref{tab:results_anim}, \ref{tab:results_ama} and \ref{tab:results_dfaust}, respectively. Note that the average values reported in the last cell in each table are computed on all the test sequences, i.e., excluding the validation sequence \texttt{cat} in ANIM, \texttt{crane} in AMA and \texttt{jumping\_jacks} in DFAUST. 

Finally, Table \ref{tab:results_stress_test} shows the results on the \texttt{horse\_collapse} sequence used for the stress test of our method, as reported in Section 4.3 of the main paper, and an additional similar sequence \texttt{camel\_collapse}. Both  sequences come from the same work of \cite{Sumner04} as the sequences \texttt{horse}, \texttt{camel} and \texttt{elephant} from the ANIM dataset, and thus we preprocess them in the same way, i.e., by scaling each sample so that the first frame of each sequence fits in a unit cube.

\begin{table}[htbp]
  \centering
  \caption{\textbf{Comparison of \ours{} to SotA methods on correspondence accuracy and reconstruction quality on the ANIM dataset}. Our method is the most accurate and also yields the same reconstruction quality as \atlasnet{}.}
  \vspace{0.1cm}
    \resizebox{0.49\textwidth}{!}{
  	\begingroup
  	\setlength{\tabcolsep}{1pt}
  	\renewcommand{\arraystretch}{1.1}
    \begin{tabular}{clcccc}
    \toprule
    \textbf{sequence} & \textbf{model} & \boldmath{}\textbf{\phantom{spac}$\mdist\downarrow$\phantom{spac}}\unboldmath{} & \boldmath{}\textbf{\phantom{spa}$\mrank\downarrow$\phantom{spa}}\unboldmath{} & \boldmath{}\textbf{\phantom{sp}$\mpckauc\uparrow$\phantom{sp}}\unboldmath{} & \boldmath{}\textbf{\phantom{spac}CD $\downarrow$\phantom{spac}}\unboldmath{} \\
    \midrule
    \multirow{5}[2]{*}{cat} & nrICP & 77.96$\pm$94.55 & 5.72$\pm$9.65 & 70.39$\pm$15.02 & - \\
          & AN    & 14.27$\pm$18.96 & 0.41$\pm$1.04 & 97.07$\pm$1.18 & \boldmath{}\textbf{0.38$\pm$0.00}\unboldmath{} \\
          & DSR   & 48.05$\pm$80.24 & 3.05$\pm$7.62 & 82.53$\pm$11.60 & 0.41$\pm$0.01 \\
          & CC    & 53.37$\pm$93.99 & 3.90$\pm$8.83 & 80.54$\pm$14.99 & - \\
          & \textbf{OUR} & \boldmath{}\textbf{9.80$\pm$14.36}\unboldmath{} & \boldmath{}\textbf{0.24$\pm$0.60}\unboldmath{} & \boldmath{}\textbf{98.27$\pm$0.82}\unboldmath{} & 0.39$\pm$0.00 \\
    \midrule
    \multirow{5}[2]{*}{horse} & nrICP & 69.94$\pm$76.34 & 4.91$\pm$7.11 & 72.62$\pm$13.23 & - \\
          & AN    & 17.52$\pm$27.33 & 0.66$\pm$3.01 & 96.60$\pm$1.24 & \boldmath{}\textbf{0.09$\pm$0.00}\unboldmath{} \\
          & DSR   & 40.21$\pm$59.73 & 2.20$\pm$4.74 & 84.31$\pm$11.61 & 0.22$\pm$0.01 \\
          & CC    & 30.24$\pm$57.97 & 1.60$\pm$4.30 & 88.39$\pm$7.95 & - \\
          & \textbf{OUR} & \boldmath{}\textbf{12.97$\pm$12.81}\unboldmath{} & \boldmath{}\textbf{0.31$\pm$0.55}\unboldmath{} & \boldmath{}\textbf{97.82$\pm$0.85}\unboldmath{} & 0.10$\pm$0.00 \\
    \midrule
    \multirow{5}[2]{*}{camel} & nrICP & 78.48$\pm$108.22 & 7.45$\pm$12.29 & 73.59$\pm$14.99 & \textbf{-} \\
          & AN    & 16.29$\pm$16.93 & 0.86$\pm$1.65 & 96.90$\pm$1.27 & 0.10$\pm$0.00 \\
          & DSR   & 75.93$\pm$116.15 & 7.43$\pm$13.56 & 73.80$\pm$13.47 & 0.17$\pm$0.02 \\
          & CC    & 56.62$\pm$93.14 & 4.75$\pm$9.32 & 77.70$\pm$14.57 & - \\
          & \textbf{OUR} & \boldmath{}\textbf{11.08$\pm$10.72}\unboldmath{} & \boldmath{}\textbf{0.42$\pm$0.83}\unboldmath{} & \boldmath{}\textbf{98.19$\pm$0.53}\unboldmath{} & \boldmath{}\textbf{0.09$\pm$0.00}\unboldmath{} \\
    \midrule
    \multirow{5}[2]{*}{elephant} & nrICP & 62.54$\pm$70.03 & 4.01$\pm$9.17 & 76.47$\pm$12.82 & - \\
          & AN    & 21.39$\pm$30.20 & 0.82$\pm$3.88 & 95.35$\pm$2.17 & 0.09$\pm$0.01 \\
          & DSR   & 23.16$\pm$26.39 & 0.68$\pm$1.82 & 93.77$\pm$3.66 & 0.19$\pm$0.00 \\
          & CC    & 14.65$\pm$11.27 & 0.27$\pm$0.63 & 97.78$\pm$0.75 & - \\
          & \textbf{OUR} & \boldmath{}\textbf{11.73$\pm$9.47}\unboldmath{} & \boldmath{}\textbf{0.16$\pm$0.33}\unboldmath{} & \boldmath{}\textbf{98.30$\pm$0.45}\unboldmath{} & \boldmath{}\textbf{0.08$\pm$0.00}\unboldmath{} \\
    \midrule
    \multirow{5}[2]{*}{MEAN} & nrICP & 70.32$\pm$84.86 & 5.46$\pm$9.52 & 74.23$\pm$13.68 & - \\
          & AN    & 18.40$\pm$24.82 & 0.78$\pm$2.85 & 96.28$\pm$1.56 & \boldmath{}\textbf{0.09$\pm$0.00}\unboldmath{} \\
          & DSR   & 46.43$\pm$67.42 & 3.44$\pm$6.71 & 83.96$\pm$9.58 & 0.19$\pm$0.01 \\
          & CC    & 33.84$\pm$54.13 & 2.21$\pm$4.75 & 87.96$\pm$7.76 & - \\
          & \textbf{OUR} & \boldmath{}\textbf{11.93$\pm$11.00}\unboldmath{} & \boldmath{}\textbf{0.30$\pm$0.57}\unboldmath{} & \boldmath{}\textbf{98.10$\pm$0.61}\unboldmath{} & \boldmath{}\textbf{0.09$\pm$0.00}\unboldmath{} \\
    \bottomrule
    \end{tabular}%
  \endgroup
	}
  \label{tab:results_anim}%
\end{table}%

\begin{table*}[htbp]
  \centering
  \caption{\textbf{Comparison of \ours{} to SotA methods on correspondence accuracy and reconstruction quality on the AMA dataset}. Our method is the most accurate and also yields reconstruction quality competitive with \atlasnet{}.}
  \vspace{0.1cm}
    \resizebox{0.99\textwidth}{!}{
  	\begingroup
  	\setlength{\tabcolsep}{1pt}
  	\renewcommand{\arraystretch}{1.1}
    \begin{tabular}{rrrrrr|clcccc}
    \toprule
    \multicolumn{1}{c}{\textbf{sequence}} & \multicolumn{1}{l}{\textbf{model}} & \multicolumn{1}{c}{\boldmath{}\textbf{\phantom{spac}$\mdist\downarrow$\phantom{spac}}\unboldmath{}} & \multicolumn{1}{c}{\boldmath{}\textbf{\phantom{spa}$\mrank\downarrow$\phantom{spa}}\unboldmath{}} & \multicolumn{1}{c}{\boldmath{}\textbf{\phantom{sp}$\mpckauc\uparrow$\phantom{sp}}\unboldmath{}} & \multicolumn{1}{c|}{\boldmath{}\textbf{\phantom{spac}CD $\downarrow$\phantom{spac}}\unboldmath{}} & \textbf{sequence} & \textbf{model} & \boldmath{}\textbf{\phantom{spac}$\mdist\downarrow$\phantom{spac}}\unboldmath{} & \boldmath{}\textbf{\phantom{spa}$\mrank\downarrow$\phantom{spa}}\unboldmath{} & \boldmath{}\textbf{\phantom{sp}$\mpckauc\uparrow$\phantom{sp}}\unboldmath{} & \boldmath{}\textbf{\phantom{spac}CD $\downarrow$\phantom{spac}}\unboldmath{} \\
    \midrule
    \multicolumn{1}{c}{\multirow{5}[2]{*}{\phantom{spacer}bouncing\phantom{spacer}}} & \multicolumn{1}{l}{nrICP} & \multicolumn{1}{c}{130.93$\pm$111.27} & \multicolumn{1}{c}{4.11$\pm$7.76} & \multicolumn{1}{c}{43.64$\pm$17.60} & \multicolumn{1}{c|}{-} & \multirow{5}[2]{*}{\phantom{spacer}march\_2\phantom{spacer}} & nrICP & 187.52$\pm$182.98 & 9.33$\pm$14.65 & 39.92$\pm$23.08 & - \\
          & \multicolumn{1}{l}{AN} & \multicolumn{1}{c}{59.86$\pm$50.14} & \multicolumn{1}{c}{0.92$\pm$3.19} & \multicolumn{1}{c}{77.84$\pm$7.56} & \multicolumn{1}{c|}{\boldmath{}\textbf{0.37$\pm$0.02}\unboldmath{}} &       & AN    & 120.96$\pm$166.77 & 5.03$\pm$11.33 & 64.99$\pm$18.34 & \boldmath{}\textbf{0.30$\pm$0.02}\unboldmath{} \\
          & \multicolumn{1}{l}{DSR} & \multicolumn{1}{c}{51.09$\pm$34.89} & \multicolumn{1}{c}{0.56$\pm$0.96} & \multicolumn{1}{c}{81.57$\pm$7.00} & \multicolumn{1}{c|}{0.35$\pm$0.01} &       & DSR   & 143.05$\pm$144.18 & 5.76$\pm$10.75 & 47.77$\pm$20.80 & 0.96$\pm$0.15 \\
          & \multicolumn{1}{l}{CC} & \multicolumn{1}{c}{45.21$\pm$28.94} & \multicolumn{1}{c}{\boldmath{}\textbf{0.40$\pm$0.60}\unboldmath{}} & \multicolumn{1}{c}{85.56$\pm$6.59} & \multicolumn{1}{c|}{-} &       & CC    & 118.02$\pm$160.88 & 5.23$\pm$11.34 & 65.18$\pm$20.49 & - \\
          & \multicolumn{1}{l}{\textbf{OUR}} & \multicolumn{1}{c}{\boldmath{}\textbf{44.31$\pm$29.47}\unboldmath{}} & \multicolumn{1}{c}{0.41$\pm$0.71} & \multicolumn{1}{c}{\boldmath{}\textbf{85.92$\pm$6.06}\unboldmath{}} & \multicolumn{1}{c|}{0.40$\pm$0.03} &       & \textbf{OUR} & \boldmath{}\textbf{90.62$\pm$156.21}\unboldmath{} & \boldmath{}\textbf{3.80$\pm$11.09}\unboldmath{} & \boldmath{}\textbf{77.99$\pm$18.38}\unboldmath{} & 0.36$\pm$0.03 \\
    \midrule
    \multicolumn{1}{c}{\multirow{5}[2]{*}{crane}} & \multicolumn{1}{l}{nrICP} & \multicolumn{1}{c}{172.55$\pm$167.76} & \multicolumn{1}{c}{7.83$\pm$12.56} & \multicolumn{1}{c}{41.61$\pm$19.29} & \multicolumn{1}{c|}{-} & \multirow{5}[2]{*}{samba} & nrICP & 100.50$\pm$91.01 & 3.58$\pm$6.17 & 58.47$\pm$23.58 & - \\
          & \multicolumn{1}{l}{AN} & \multicolumn{1}{c}{77.64$\pm$99.96} & \multicolumn{1}{c}{2.13$\pm$6.27} & \multicolumn{1}{c}{74.08$\pm$12.38} & \multicolumn{1}{c|}{\boldmath{}\textbf{0.30$\pm$0.01}\unboldmath{}} &       & AN    & 75.38$\pm$83.78 & 2.41$\pm$5.42 & 72.22$\pm$20.14 & \boldmath{}\textbf{0.22$\pm$0.01}\unboldmath{} \\
          & \multicolumn{1}{l}{DSR} & \multicolumn{1}{c}{128.18$\pm$168.36} & \multicolumn{1}{c}{5.64$\pm$11.18} & \multicolumn{1}{c}{64.07$\pm$23.11} & \multicolumn{1}{c|}{0.29$\pm$0.01} &       & DSR   & 114.61$\pm$124.73 & 5.06$\pm$9.48 & 57.38$\pm$26.40 & 0.25$\pm$0.01 \\
          & \multicolumn{1}{l}{CC} & \multicolumn{1}{c}{76.80$\pm$96.52} & \multicolumn{1}{c}{2.17$\pm$5.85} & \multicolumn{1}{c}{73.49$\pm$18.45} & \multicolumn{1}{c|}{-} &       & CC    & 72.31$\pm$85.51 & 2.34$\pm$5.49 & 73.72$\pm$20.53 & - \\
          & \multicolumn{1}{l}{\textbf{OUR}} & \multicolumn{1}{c}{\boldmath{}\textbf{66.63$\pm$103.11}\unboldmath{}} & \multicolumn{1}{c}{\boldmath{}\textbf{2.11$\pm$6.75}\unboldmath{}} & \multicolumn{1}{c}{\boldmath{}\textbf{80.24$\pm$11.42}\unboldmath{}} & \multicolumn{1}{c|}{0.31$\pm$0.02} &       & \textbf{OUR} & \boldmath{}\textbf{62.84$\pm$80.00}\unboldmath{} & \boldmath{}\textbf{1.93$\pm$4.64}\unboldmath{} & \boldmath{}\textbf{77.44$\pm$21.38}\unboldmath{} & 0.27$\pm$0.02 \\
    \midrule
    \multicolumn{1}{c}{\multirow{5}[2]{*}{handstand}} & \multicolumn{1}{l}{nrICP} & \multicolumn{1}{c}{256.22$\pm$194.98} & \multicolumn{1}{c}{14.16$\pm$17.45} & \multicolumn{1}{c}{21.87$\pm$24.70} & \multicolumn{1}{c|}{-} & \multirow{5}[2]{*}{squat\_1} & nrICP & 84.88$\pm$86.77 & 2.23$\pm$4.93 & 65.16$\pm$26.07 & - \\
          & \multicolumn{1}{l}{AN} & \multicolumn{1}{c}{182.36$\pm$186.97} & \multicolumn{1}{c}{8.74$\pm$14.38} & \multicolumn{1}{c}{42.11$\pm$26.55} & \multicolumn{1}{c|}{\boldmath{}\textbf{0.35$\pm$0.03}\unboldmath{}} &       & AN    & 46.91$\pm$41.54 & 0.71$\pm$1.59 & 82.91$\pm$12.40 & 0.28$\pm$0.01 \\
          & \multicolumn{1}{l}{DSR} & \multicolumn{1}{c}{380.79$\pm$226.71} & \multicolumn{1}{c}{20.52$\pm$17.35} & \multicolumn{1}{c}{4.54$\pm$1.59} & \multicolumn{1}{c|}{551.20$\pm$471.57} &       & DSR   & 46.91$\pm$41.82 & 0.65$\pm$1.52 & 82.67$\pm$13.36 & 0.28$\pm$0.01 \\
          & \multicolumn{1}{l}{CC} & \multicolumn{1}{c}{\boldmath{}\textbf{89.79$\pm$142.39}\unboldmath{}} & \multicolumn{1}{c}{\boldmath{}\textbf{3.11$\pm$10.10}\unboldmath{}} & \multicolumn{1}{c}{\boldmath{}\textbf{71.53$\pm$16.92}\unboldmath{}} & \multicolumn{1}{c|}{-} &       & CC    & \boldmath{}\textbf{26.81$\pm$18.42}\unboldmath{} & \boldmath{}\textbf{0.16$\pm$0.25}\unboldmath{} & \boldmath{}\textbf{94.32$\pm$2.83}\unboldmath{} & - \\
          & \multicolumn{1}{l}{\textbf{OUR}} & \multicolumn{1}{c}{126.52$\pm$167.36} & \multicolumn{1}{c}{5.69$\pm$13.80} & \multicolumn{1}{c}{58.47$\pm$23.49} & \multicolumn{1}{c|}{0.38$\pm$0.03} &       & \textbf{OUR} & 27.81$\pm$27.48 & 0.25$\pm$0.78 & 92.60$\pm$4.73 & \boldmath{}\textbf{0.27$\pm$0.00}\unboldmath{} \\
    \midrule
    \multicolumn{1}{c}{\multirow{5}[2]{*}{jumping}} & \multicolumn{1}{l}{nrICP} & \multicolumn{1}{c}{206.43$\pm$172.56} & \multicolumn{1}{c}{10.38$\pm$14.50} & \multicolumn{1}{c}{29.02$\pm$18.96} & \multicolumn{1}{c|}{-} & \multirow{5}[2]{*}{squat\_2} & nrICP & 90.50$\pm$85.60 & 2.29$\pm$4.77 & 61.18$\pm$25.81 & - \\
          & \multicolumn{1}{l}{AN} & \multicolumn{1}{c}{116.86$\pm$148.11} & \multicolumn{1}{c}{4.86$\pm$11.78} & \multicolumn{1}{c}{61.05$\pm$18.89} & \multicolumn{1}{c|}{\boldmath{}\textbf{0.32$\pm$0.02}\unboldmath{}} &       & AN    & 47.93$\pm$42.15 & 0.66$\pm$1.63 & 82.61$\pm$10.97 & 0.29$\pm$0.01 \\
          & \multicolumn{1}{l}{DSR} & \multicolumn{1}{c}{114.90$\pm$151.03} & \multicolumn{1}{c}{4.92$\pm$11.16} & \multicolumn{1}{c}{64.81$\pm$19.44} & \multicolumn{1}{c|}{0.33$\pm$0.02} &       & DSR   & 121.50$\pm$119.10 & 3.75$\pm$6.51 & 51.09$\pm$27.88 & 4.77$\pm$0.71 \\
          & \multicolumn{1}{l}{CC} & \multicolumn{1}{c}{77.53$\pm$111.13} & \multicolumn{1}{c}{2.29$\pm$6.71} & \multicolumn{1}{c}{75.31$\pm$17.45} & \multicolumn{1}{c|}{-} &       & CC    & 37.02$\pm$28.14 & \boldmath{}\textbf{0.32$\pm$0.58}\unboldmath{} & 89.14$\pm$7.37 & - \\
          & \multicolumn{1}{l}{\textbf{OUR}} & \multicolumn{1}{c}{\boldmath{}\textbf{45.10$\pm$31.25}\unboldmath{}} & \multicolumn{1}{c}{\boldmath{}\textbf{0.50$\pm$1.01}\unboldmath{}} & \multicolumn{1}{c}{\boldmath{}\textbf{85.31$\pm$6.01}\unboldmath{}} & \multicolumn{1}{c|}{0.37$\pm$0.03} &       & \textbf{OUR} & \boldmath{}\textbf{33.72$\pm$31.41}\unboldmath{} & \boldmath{}\textbf{0.32$\pm$0.81}\unboldmath{} & \boldmath{}\textbf{89.85$\pm$7.06}\unboldmath{} & \boldmath{}\textbf{0.28$\pm$0.01}\unboldmath{} \\
    \midrule
    \multicolumn{1}{c}{\multirow{5}[2]{*}{march\_1}} & \multicolumn{1}{l}{nrICP} & \multicolumn{1}{c}{174.05$\pm$171.86} & \multicolumn{1}{c}{8.59$\pm$14.23} & \multicolumn{1}{c}{42.75$\pm$22.83} & \multicolumn{1}{c|}{-} & \multirow{5}[2]{*}{swing} & nrICP & 127.41$\pm$111.79 & 4.98$\pm$7.89 & 46.58$\pm$17.80 & - \\
          & \multicolumn{1}{l}{AN} & \multicolumn{1}{c}{57.66$\pm$42.75} & \multicolumn{1}{c}{0.93$\pm$1.99} & \multicolumn{1}{c}{77.90$\pm$9.56} & \multicolumn{1}{c|}{\boldmath{}\textbf{0.29$\pm$0.01}\unboldmath{}} &       & AN    & 73.26$\pm$59.32 & 1.86$\pm$4.30 & 69.01$\pm$13.40 & \boldmath{}\textbf{0.24$\pm$0.02}\unboldmath{} \\
          & \multicolumn{1}{l}{DSR} & \multicolumn{1}{c}{79.26$\pm$97.42} & \multicolumn{1}{c}{2.52$\pm$6.44} & \multicolumn{1}{c}{72.20$\pm$14.13} & \multicolumn{1}{c|}{0.33$\pm$0.02} &       & DSR   & 59.92$\pm$49.39 & 1.25$\pm$2.38 & 75.19$\pm$12.90 & \boldmath{}\textbf{0.24$\pm$0.01}\unboldmath{} \\
          & \multicolumn{1}{l}{CC} & \multicolumn{1}{c}{124.76$\pm$157.34} & \multicolumn{1}{c}{5.26$\pm$10.01} & \multicolumn{1}{c}{64.16$\pm$23.63} & \multicolumn{1}{c|}{-} &       & CC    & 79.78$\pm$149.09 & 3.13$\pm$12.27 & 74.73$\pm$19.23 & - \\
          & \multicolumn{1}{l}{\textbf{OUR}} & \multicolumn{1}{c}{\boldmath{}\textbf{34.85$\pm$24.72}\unboldmath{}} & \multicolumn{1}{c}{\boldmath{}\textbf{0.29$\pm$0.56}\unboldmath{}} & \multicolumn{1}{c}{\boldmath{}\textbf{90.66$\pm$4.66}\unboldmath{}} & \multicolumn{1}{c|}{0.31$\pm$0.01} &       & \textbf{OUR} & \boldmath{}\textbf{48.28$\pm$40.07}\unboldmath{} & \boldmath{}\textbf{0.80$\pm$1.67}\unboldmath{} & \boldmath{}\textbf{82.39$\pm$8.68}\unboldmath{} & 0.26$\pm$0.01 \\
    \midrule
          &       &       &       &       &       & \multirow{5}[2]{*}{MEAN} & nrICP & 150.94$\pm$134.31 & 6.63$\pm$10.26 & 45.40$\pm$22.27 & - \\
          &       &       &       &       &       &       & AN    & 86.80$\pm$91.28 & 2.90$\pm$6.18 & 70.07$\pm$15.31 & \boldmath{}\textbf{0.30$\pm$0.01}\unboldmath{} \\
          &       &       &       &       &       &       & DSR   & 123.56$\pm$109.92 & 5.00$\pm$7.39 & 59.69$\pm$15.94 & 62.08$\pm$52.50 \\
          &       &       &       &       &       &       & CC    & 74.58$\pm$97.98 & 2.47$\pm$6.37 & 77.07$\pm$15.00 & - \\
          &       &       &       &       &       &       & \textbf{OUR} & \boldmath{}\textbf{57.12$\pm$65.33}\unboldmath{} & \boldmath{}\textbf{1.55$\pm$3.90}\unboldmath{} & \boldmath{}\textbf{82.29$\pm$11.16}\unboldmath{} & 0.32$\pm$0.02 \\
\cmidrule{7-12}    \end{tabular}%
  \endgroup
	}
  \label{tab:results_ama}%
\end{table*}%

\begin{table*}[htbp]
  \centering
  \caption{\textbf{Comparison of \ours{} to SotA methods on correspondence accuracy and reconstruction quality on the DFAUST dataset}. Our method is the most accurate and also yields reconstruction quality on par with \atlasnet{}.}
  \vspace{0.1cm}
    \resizebox{0.99\textwidth}{!}{
  	\begingroup
  	\setlength{\tabcolsep}{1pt}
  	\renewcommand{\arraystretch}{1.1}
    \begin{tabular}{rrrrrr|clcccc}
    \toprule
    \multicolumn{1}{c}{\textbf{sequence}} & \multicolumn{1}{l}{\textbf{model}} & \multicolumn{1}{c}{\boldmath{}\textbf{\phantom{spac}$\mdist\downarrow$\phantom{spac}}\unboldmath{}} & \multicolumn{1}{c}{\boldmath{}\textbf{\phantom{spa}$\mrank\downarrow$\phantom{spa}}\unboldmath{}} & \multicolumn{1}{c}{\boldmath{}\textbf{\phantom{sp}$\mpckauc\uparrow$\phantom{sp}}\unboldmath{}} & \multicolumn{1}{c|}{\boldmath{}\textbf{\phantom{spac}CD $\downarrow$\phantom{spac}}\unboldmath{}} & \textbf{sequence} & \textbf{model} & \boldmath{}\textbf{\phantom{spac}$\mdist\downarrow$\phantom{spac}}\unboldmath{} & \boldmath{}\textbf{\phantom{spa}$\mrank\downarrow$\phantom{spa}}\unboldmath{} & \boldmath{}\textbf{\phantom{sp}$\mpckauc\uparrow$\phantom{sp}}\unboldmath{} & \boldmath{}\textbf{\phantom{spac}CD $\downarrow$\phantom{spac}}\unboldmath{} \\
    \midrule
    \multicolumn{1}{c}{\multirow{5}[2]{*}{chicken\_wings}} & \multicolumn{1}{l}{nrICP} & \multicolumn{1}{c}{72.20$\pm$135.09} & \multicolumn{1}{c}{5.15$\pm$15.70} & \multicolumn{1}{c}{80.89$\pm$13.79} & \multicolumn{1}{c|}{-} & \multirow{5}[2]{*}{one\_leg\_jump} & nrICP & 81.13$\pm$98.03 & 2.86$\pm$5.19 & 71.68$\pm$14.61 & - \\
          & \multicolumn{1}{l}{AN} & \multicolumn{1}{c}{49.75$\pm$92.05} & \multicolumn{1}{c}{2.31$\pm$6.96} & \multicolumn{1}{c}{85.57$\pm$16.07} & \multicolumn{1}{c|}{\boldmath{}\textbf{0.37$\pm$0.08}\unboldmath{}} &       & AN    & 30.65$\pm$23.71 & 0.58$\pm$0.98 & \boldmath{}\textbf{92.47$\pm$3.48}\unboldmath{} & 0.43$\pm$0.08 \\
          & \multicolumn{1}{l}{DSR} & \multicolumn{1}{c}{282.61$\pm$130.70} & \multicolumn{1}{c}{21.85$\pm$17.43} & \multicolumn{1}{c}{6.40$\pm$2.66} & \multicolumn{1}{c|}{97.10$\pm$25.04} &       & DSR   & 47.59$\pm$99.20 & 1.30$\pm$4.72 & 88.44$\pm$8.96 & \boldmath{}\textbf{0.36$\pm$0.07}\unboldmath{} \\
          & \multicolumn{1}{l}{CC} & \multicolumn{1}{c}{35.57$\pm$113.74} & \multicolumn{1}{c}{2.12$\pm$11.40} & \multicolumn{1}{c}{93.59$\pm$16.90} & \multicolumn{1}{c|}{-} &       & CC    & \boldmath{}\textbf{36.40$\pm$76.87}\unboldmath{} & \boldmath{}\textbf{0.79$\pm$3.47}\unboldmath{} & 91.23$\pm$8.64 & - \\
          & \multicolumn{1}{l}{\textbf{OUR}} & \multicolumn{1}{c}{\boldmath{}\textbf{18.51$\pm$20.37}\unboldmath{}} & \multicolumn{1}{c}{\boldmath{}\textbf{0.38$\pm$1.34}\unboldmath{}} & \multicolumn{1}{c}{\boldmath{}\textbf{96.53$\pm$1.98}\unboldmath{}} & \multicolumn{1}{c|}{0.43$\pm$0.09} &       & \textbf{OUR} & 41.44$\pm$74.93 & 1.03$\pm$3.56 & 87.92$\pm$11.02 & \boldmath{}\textbf{0.36$\pm$0.07}\unboldmath{} \\
    \midrule
    \multicolumn{1}{c}{\multirow{5}[2]{*}{hips}} & \multicolumn{1}{l}{nrICP} & \multicolumn{1}{c}{59.22$\pm$49.65} & \multicolumn{1}{c}{2.39$\pm$3.91} & \multicolumn{1}{c}{76.95$\pm$14.85} & \multicolumn{1}{c|}{-} & \multirow{5}[2]{*}{one\_leg\_loose} & nrICP & 54.98$\pm$64.48 & 1.74$\pm$3.31 & 81.73$\pm$11.45 & - \\
          & \multicolumn{1}{l}{AN} & \multicolumn{1}{c}{21.16$\pm$16.45} & \multicolumn{1}{c}{0.33$\pm$0.60} & \multicolumn{1}{c}{96.06$\pm$1.58} & \multicolumn{1}{c|}{0.27$\pm$0.05} &       & AN    & 31.92$\pm$35.99 & 0.68$\pm$1.77 & 91.35$\pm$6.00 & \boldmath{}\textbf{0.38$\pm$0.07}\unboldmath{} \\
          & \multicolumn{1}{l}{DSR} & \multicolumn{1}{c}{17.33$\pm$14.65} & \multicolumn{1}{c}{0.29$\pm$0.62} & \multicolumn{1}{c}{97.11$\pm$1.10} & \multicolumn{1}{c|}{0.28$\pm$0.04} &       & DSR   & 23.53$\pm$24.66 & 0.41$\pm$0.80 & 94.93$\pm$3.59 & 0.43$\pm$0.09 \\
          & \multicolumn{1}{l}{CC} & \multicolumn{1}{c}{41.57$\pm$138.23} & \multicolumn{1}{c}{2.19$\pm$12.00} & \multicolumn{1}{c}{93.30$\pm$16.63} & \multicolumn{1}{c|}{-} &       & CC    & 23.90$\pm$31.93 & 0.51$\pm$1.85 & 94.61$\pm$11.02 & - \\
          & \multicolumn{1}{l}{\textbf{OUR}} & \multicolumn{1}{c}{\boldmath{}\textbf{14.76$\pm$12.21}\unboldmath{}} & \multicolumn{1}{c}{\boldmath{}\textbf{0.23$\pm$0.51}\unboldmath{}} & \multicolumn{1}{c}{\boldmath{}\textbf{97.83$\pm$0.77}\unboldmath{}} & \multicolumn{1}{c|}{\boldmath{}\textbf{0.25$\pm$0.04}\unboldmath{}} &       & \textbf{OUR} & \boldmath{}\textbf{18.40$\pm$15.28}\unboldmath{} & \boldmath{}\textbf{0.29$\pm$0.53}\unboldmath{} & \boldmath{}\textbf{96.92$\pm$1.61}\unboldmath{} & 0.41$\pm$0.07 \\
    \midrule
    \multicolumn{1}{c}{\multirow{5}[2]{*}{jiggle\_on\_toes}} & \multicolumn{1}{l}{nrICP} & \multicolumn{1}{c}{128.60$\pm$194.22} & \multicolumn{1}{c}{7.46$\pm$16.20} & \multicolumn{1}{c}{59.47$\pm$25.80} & \multicolumn{1}{c|}{-} & \multirow{5}[2]{*}{punching} & nrICP & 117.92$\pm$175.81 & 8.29$\pm$18.23 & 65.40$\pm$19.74 & - \\
          & \multicolumn{1}{l}{AN} & \multicolumn{1}{c}{35.43$\pm$72.20} & \multicolumn{1}{c}{1.34$\pm$5.59} & \multicolumn{1}{c}{90.04$\pm$9.74} & \multicolumn{1}{c|}{0.31$\pm$0.05} &       & AN    & 43.01$\pm$64.98 & 1.70$\pm$6.07 & 87.19$\pm$9.49 & 0.36$\pm$0.06 \\
          & \multicolumn{1}{l}{DSR} & \multicolumn{1}{c}{29.01$\pm$47.00} & \multicolumn{1}{c}{0.95$\pm$3.74} & \multicolumn{1}{c}{92.70$\pm$7.63} & \multicolumn{1}{c|}{0.38$\pm$0.05} &       & DSR   & 204.86$\pm$141.43 & 13.04$\pm$15.68 & 15.99$\pm$4.52 & 39.66$\pm$12.26 \\
          & \multicolumn{1}{l}{CC} & \multicolumn{1}{c}{26.26$\pm$69.02} & \multicolumn{1}{c}{0.91$\pm$5.71} & \multicolumn{1}{c}{94.86$\pm$10.98} & \multicolumn{1}{c|}{-} &       & CC    & \boldmath{}\textbf{21.13$\pm$17.43}\unboldmath{} & \boldmath{}\textbf{0.41$\pm$0.89}\unboldmath{} & \boldmath{}\textbf{95.94$\pm$2.14}\unboldmath{} & - \\
          & \multicolumn{1}{l}{\textbf{OUR}} & \multicolumn{1}{c}{\boldmath{}\textbf{16.45$\pm$14.24}\unboldmath{}} & \multicolumn{1}{c}{\boldmath{}\textbf{0.28$\pm$0.69}\unboldmath{}} & \multicolumn{1}{c}{\boldmath{}\textbf{97.38$\pm$1.31}\unboldmath{}} & \multicolumn{1}{c|}{\boldmath{}\textbf{0.30$\pm$0.05}\unboldmath{}} &       & \textbf{OUR} & 21.88$\pm$24.82 & 0.50$\pm$1.44 & 95.51$\pm$2.90 & \boldmath{}\textbf{0.32$\pm$0.06}\unboldmath{} \\
    \midrule
    \multicolumn{1}{c}{\multirow{5}[2]{*}{jumping\ jacks}} & \multicolumn{1}{l}{nrICP} & \multicolumn{1}{c}{187.37$\pm$240.34} & \multicolumn{1}{c}{11.78$\pm$20.47} & \multicolumn{1}{c}{46.37$\pm$28.06} & \multicolumn{1}{c|}{-} & \multirow{5}[2]{*}{running\_on\_spot} & nrICP & 103.57$\pm$129.95 & 5.57$\pm$12.83 & 65.36$\pm$15.24 & - \\
          & \multicolumn{1}{l}{AN} & \multicolumn{1}{c}{51.76$\pm$60.73} & \multicolumn{1}{c}{1.95$\pm$4.67} & \multicolumn{1}{c}{81.68$\pm$10.34} & \multicolumn{1}{c|}{0.52$\pm$0.08} &       & AN    & 39.77$\pm$46.21 & 1.18$\pm$4.33 & 88.93$\pm$4.76 & 0.49$\pm$0.09 \\
          & \multicolumn{1}{l}{DSR} & \multicolumn{1}{c}{35.25$\pm$55.18} & \multicolumn{1}{c}{1.07$\pm$4.01} & \multicolumn{1}{c}{90.60$\pm$5.07} & \multicolumn{1}{c|}{0.43$\pm$0.05} &       & DSR   & 43.93$\pm$58.32 & 1.57$\pm$4.44 & 86.86$\pm$8.58 & 0.45$\pm$0.05 \\
          & \multicolumn{1}{l}{CC} & \multicolumn{1}{c}{32.74$\pm$31.65} & \multicolumn{1}{c}{0.70$\pm$1.68} & \multicolumn{1}{c}{91.47$\pm$6.25} & \multicolumn{1}{c|}{-} &       & CC    & 26.64$\pm$20.92 & 0.52$\pm$1.25 & 94.38$\pm$3.04 & - \\
          & \multicolumn{1}{l}{\textbf{OUR}} & \multicolumn{1}{c}{\boldmath{}\textbf{27.98$\pm$38.15}\unboldmath{}} & \multicolumn{1}{c}{\boldmath{}\textbf{0.67$\pm$2.55}\unboldmath{}} & \multicolumn{1}{c}{\boldmath{}\textbf{93.65$\pm$3.15}\unboldmath{}} & \multicolumn{1}{c|}{\boldmath{}\textbf{0.41$\pm$0.08}\unboldmath{}} &       & \textbf{OUR} & \boldmath{}\textbf{21.28$\pm$18.99}\unboldmath{} & \boldmath{}\textbf{0.39$\pm$1.39}\unboldmath{} & \boldmath{}\textbf{96.07$\pm$1.64}\unboldmath{} & \boldmath{}\textbf{0.32$\pm$0.05}\unboldmath{} \\
    \midrule
    \multicolumn{1}{c}{\multirow{5}[2]{*}{knees}} & \multicolumn{1}{l}{nrICP} & \multicolumn{1}{c}{116.26$\pm$157.30} & \multicolumn{1}{c}{5.22$\pm$10.39} & \multicolumn{1}{c}{61.25$\pm$20.85} & \multicolumn{1}{c|}{-} & \multirow{5}[2]{*}{shake\_arms} & nrICP & 88.38$\pm$158.81 & 5.40$\pm$14.90 & 75.74$\pm$18.18 & - \\
          & \multicolumn{1}{l}{AN} & \multicolumn{1}{c}{43.33$\pm$81.94} & \multicolumn{1}{c}{1.02$\pm$3.72} & \multicolumn{1}{c}{89.00$\pm$5.32} & \multicolumn{1}{c|}{\boldmath{}\textbf{0.40$\pm$0.09}\unboldmath{}} &       & AN    & 28.13$\pm$31.89 & 0.81$\pm$2.35 & 92.92$\pm$4.44 & \boldmath{}\textbf{0.34$\pm$0.06}\unboldmath{} \\
          & \multicolumn{1}{l}{DSR} & \multicolumn{1}{c}{24.76$\pm$22.94} & \multicolumn{1}{c}{0.43$\pm$0.79} & \multicolumn{1}{c}{94.25$\pm$2.69} & \multicolumn{1}{c|}{0.47$\pm$0.11} &       & DSR   & 25.99$\pm$27.03 & 0.77$\pm$1.85 & 93.67$\pm$3.30 & 0.50$\pm$0.12 \\
          & \multicolumn{1}{l}{CC} & \multicolumn{1}{c}{30.38$\pm$86.52} & \multicolumn{1}{c}{1.14$\pm$7.12} & \multicolumn{1}{c}{94.47$\pm$13.35} & \multicolumn{1}{c|}{-} &       & CC    & 21.82$\pm$18.50 & 0.51$\pm$1.02 & 95.61$\pm$2.01 & - \\
          & \multicolumn{1}{l}{\textbf{OUR}} & \multicolumn{1}{c}{\boldmath{}\textbf{23.45$\pm$19.49}\unboldmath{}} & \multicolumn{1}{c}{\boldmath{}\textbf{0.40$\pm$0.74}\unboldmath{}} & \multicolumn{1}{c}{\boldmath{}\textbf{95.23$\pm$2.37}\unboldmath{}} & \multicolumn{1}{c|}{0.52$\pm$0.12} &       & \textbf{OUR} & \boldmath{}\textbf{17.29$\pm$19.21}\unboldmath{} & \boldmath{}\textbf{0.41$\pm$1.34}\unboldmath{} & \boldmath{}\textbf{96.97$\pm$1.31}\unboldmath{} & 0.36$\pm$0.08 \\
    \midrule
    \multicolumn{1}{c}{\multirow{5}[2]{*}{light\_hopping\_loose}} & \multicolumn{1}{l}{nrICP} & \multicolumn{1}{c}{41.15$\pm$29.98} & \multicolumn{1}{c}{1.29$\pm$2.24} & \multicolumn{1}{c}{87.08$\pm$6.84} & \multicolumn{1}{c|}{-} & \multirow{5}[2]{*}{shake\_hips} & nrICP & 86.76$\pm$156.94 & 5.00$\pm$14.35 & 76.47$\pm$17.42 & - \\
          & \multicolumn{1}{l}{AN} & \multicolumn{1}{c}{21.46$\pm$21.06} & \multicolumn{1}{c}{0.39$\pm$1.14} & \multicolumn{1}{c}{95.83$\pm$2.04} & \multicolumn{1}{c|}{\boldmath{}\textbf{0.28$\pm$0.05}\unboldmath{}} &       & AN    & 28.27$\pm$28.74 & 0.71$\pm$1.77 & 92.46$\pm$5.48 & 0.29$\pm$0.05 \\
          & \multicolumn{1}{l}{DSR} & \multicolumn{1}{c}{21.81$\pm$20.35} & \multicolumn{1}{c}{0.51$\pm$1.18} & \multicolumn{1}{c}{95.48$\pm$2.91} & \multicolumn{1}{c|}{0.40$\pm$0.05} &       & DSR   & 92.23$\pm$158.27 & 5.22$\pm$13.78 & 75.27$\pm$19.64 & 1.28$\pm$0.37 \\
          & \multicolumn{1}{l}{CC} & \multicolumn{1}{c}{25.61$\pm$53.71} & \multicolumn{1}{c}{0.82$\pm$4.13} & \multicolumn{1}{c}{94.66$\pm$12.36} & \multicolumn{1}{c|}{-} &       & CC    & 49.93$\pm$160.68 & 3.21$\pm$15.05 & 91.60$\pm$19.67 & - \\
          & \multicolumn{1}{l}{\textbf{OUR}} & \multicolumn{1}{c}{\boldmath{}\textbf{16.12$\pm$12.56}\unboldmath{}} & \multicolumn{1}{c}{\boldmath{}\textbf{0.27$\pm$0.54}\unboldmath{}} & \multicolumn{1}{c}{\boldmath{}\textbf{97.56$\pm$0.96}\unboldmath{}} & \multicolumn{1}{c|}{0.32$\pm$0.06} &       & \textbf{OUR} & \boldmath{}\textbf{17.60$\pm$16.05}\unboldmath{} & \boldmath{}\textbf{0.29$\pm$0.89}\unboldmath{} & \boldmath{}\textbf{96.90$\pm$1.73}\unboldmath{} & \boldmath{}\textbf{0.28$\pm$0.05}\unboldmath{} \\
    \midrule
    \multicolumn{1}{c}{\multirow{5}[2]{*}{light\_hopping\_stiff}} & \multicolumn{1}{l}{nrICP} & \multicolumn{1}{c}{33.16$\pm$24.91} & \multicolumn{1}{c}{0.93$\pm$1.80} & \multicolumn{1}{c}{91.11$\pm$3.93} & \multicolumn{1}{c|}{-} & \multirow{5}[2]{*}{shake\_shoulders} & nrICP & 53.85$\pm$42.97 & 1.90$\pm$2.84 & 79.19$\pm$11.83 & - \\
          & \multicolumn{1}{l}{AN} & \multicolumn{1}{c}{17.30$\pm$15.16} & \multicolumn{1}{c}{0.25$\pm$0.62} & \multicolumn{1}{c}{97.05$\pm$1.25} & \multicolumn{1}{c|}{0.28$\pm$0.05} &       & AN    & 22.48$\pm$17.33 & 0.39$\pm$0.68 & 95.63$\pm$1.78 & \boldmath{}\textbf{0.29$\pm$0.05}\unboldmath{} \\
          & \multicolumn{1}{l}{DSR} & \multicolumn{1}{c}{58.32$\pm$36.51} & \multicolumn{1}{c}{2.05$\pm$2.81} & \multicolumn{1}{c}{77.38$\pm$14.48} & \multicolumn{1}{c|}{4.11$\pm$2.07} &       & DSR   & 22.27$\pm$18.30 & 0.43$\pm$0.78 & 95.48$\pm$2.30 & 0.31$\pm$0.04 \\
          & \multicolumn{1}{l}{CC} & \multicolumn{1}{c}{25.50$\pm$79.93} & \multicolumn{1}{c}{1.14$\pm$7.54} & \multicolumn{1}{c}{95.63$\pm$13.47} & \multicolumn{1}{c|}{-} &       & CC    & 19.67$\pm$14.47 & \boldmath{}\textbf{0.31$\pm$0.51}\unboldmath{} & 96.63$\pm$1.08 & - \\
          & \multicolumn{1}{l}{\textbf{OUR}} & \multicolumn{1}{c}{\boldmath{}\textbf{12.21$\pm$9.95}\unboldmath{}} & \multicolumn{1}{c}{\boldmath{}\textbf{0.16$\pm$0.30}\unboldmath{}} & \multicolumn{1}{c}{\boldmath{}\textbf{98.40$\pm$0.39}\unboldmath{}} & \multicolumn{1}{c|}{\boldmath{}\textbf{0.27$\pm$0.05}\unboldmath{}} &       & \textbf{OUR} & \boldmath{}\textbf{18.08$\pm$14.37}\unboldmath{} & 0.32$\pm$0.58 & \boldmath{}\textbf{96.97$\pm$1.19}\unboldmath{} & 0.30$\pm$0.05 \\
    \midrule
          &       &       &       &       &       & \multirow{5}[2]{*}{MEAN} & nrICP & 79.78$\pm$118.46 & 4.09$\pm$10.17 & 74.79$\pm$15.90 & - \\
          &       &       &       &       &       &       & AN    & 31.74$\pm$43.46 & 0.90$\pm$2.95 & 91.88$\pm$5.84 & \boldmath{}\textbf{0.34$\pm$0.06}\unboldmath{} \\
          &       &       &       &       &       &       & DSR   & 68.79$\pm$61.04 & 3.76$\pm$5.19 & 78.00$\pm$6.25 & 11.21$\pm$2.89 \\
          &       &       &       &       &       &       & CC    & 29.57$\pm$65.26 & 1.12$\pm$5.26 & 94.35$\pm$9.82 & - \\
          &       &       &       &       &       &       & \textbf{OUR} & \boldmath{}\textbf{19.81$\pm$22.19}\unboldmath{} & \boldmath{}\textbf{0.38$\pm$1.17}\unboldmath{} & \boldmath{}\textbf{96.17$\pm$2.31}\unboldmath{} & \boldmath{}\textbf{0.34$\pm$0.06}\unboldmath{} \\
\cmidrule{7-12}    \end{tabular}%
  \endgroup
	}
  \label{tab:results_dfaust}%
\end{table*}%

\begin{table}[htbp]
  \centering
  \caption{\textbf{Comparison of \ours{} to SotA methods on correspondence accuracy and reconstruction quality on a collapsing rubber horse used to stress test our method and on an additional similar sequence depicting a collapsing camel}. Our method is the most accurate and also yields the same reconstruction quality as \atlasnet{}.}
  \vspace{0.1cm}
    \resizebox{0.49\textwidth}{!}{
  	\begingroup
  	\setlength{\tabcolsep}{1pt}
  	\renewcommand{\arraystretch}{1.1}
    \begin{tabular}{clcccc}
    \toprule
    \textbf{sequence} & \textbf{model} & \boldmath{}\textbf{\phantom{spac}$\mdist\downarrow$\phantom{spac}}\unboldmath{} & \boldmath{}\textbf{\phantom{spa}$\mrank\downarrow$\phantom{spa}}\unboldmath{} & \boldmath{}\textbf{\phantom{sp}$\mpckauc\uparrow$\phantom{sp}}\unboldmath{} & \boldmath{}\textbf{\phantom{spac}CD $\downarrow$\phantom{spac}}\unboldmath{} \\
    \midrule
    \multirow{5}[2]{*}{horse\_collapse} & nrICP & 54.32$\pm$46.49 & 3.88$\pm$5.47 & 78.36$\pm$13.66 & - \\
          & AN    & 62.38$\pm$79.58 & 4.91$\pm$9.05 & 74.81$\pm$18.23 & \boldmath{}\textbf{0.13$\pm$0.01}\unboldmath{} \\
          & DSR   & 49.00$\pm$60.40 & 3.25$\pm$6.18 & 81.73$\pm$14.42 & 0.18$\pm$0.02 \\
          & CC    & 56.51$\pm$78.80 & 3.89$\pm$7.49 & 77.97$\pm$19.67 & - \\
          & \textbf{OUR} & \boldmath{}\textbf{23.82$\pm$39.39}\unboldmath{} & \boldmath{}\textbf{1.11$\pm$3.48}\unboldmath{} & \boldmath{}\textbf{93.32$\pm$6.48}\unboldmath{} & \boldmath{}\textbf{0.13$\pm$0.01}\unboldmath{} \\
    \midrule
    \multirow{5}[2]{*}{camel\_collapse} & nrICP & 40.68$\pm$36.05 & 2.76$\pm$3.69 & 86.60$\pm$9.77 & - \\
          & AN    & 43.78$\pm$61.53 & 3.05$\pm$6.11 & 85.05$\pm$12.47 & 0.16$\pm$0.01 \\
          & DSR   & 67.16$\pm$96.21 & 5.12$\pm$8.99 & 75.66$\pm$19.57 & 0.25$\pm$0.02 \\
          & CC    & 349.08$\pm$371.70 & 33.96$\pm$37.24 & 48.85$\pm$48.25 & - \\
          & \textbf{OUR} & \boldmath{}\textbf{19.25$\pm$28.05}\unboldmath{} & \boldmath{}\textbf{0.81$\pm$2.09}\unboldmath{} & \boldmath{}\textbf{95.72$\pm$3.89}\unboldmath{} & \boldmath{}\textbf{0.15$\pm$0.01}\unboldmath{} \\
    \midrule
    \multirow{5}[2]{*}{MEAN} & nrICP & 47.50$\pm$41.27 & 3.32$\pm$4.58 & 82.48$\pm$11.72 & - \\
          & AN    & 53.08$\pm$70.56 & 3.98$\pm$7.58 & 79.93$\pm$15.35 & \boldmath{}\textbf{0.14$\pm$0.01}\unboldmath{} \\
          & DSR   & 58.08$\pm$78.31 & 4.19$\pm$7.59 & 78.70$\pm$17.00 & 0.21$\pm$0.02 \\
          & CC    & 202.80$\pm$225.25 & 18.93$\pm$22.37 & 63.41$\pm$33.96 & - \\
          & \textbf{OUR} & \boldmath{}\textbf{21.54$\pm$33.72}\unboldmath{} & \boldmath{}\textbf{0.96$\pm$2.79}\unboldmath{} & \boldmath{}\textbf{94.52$\pm$5.19}\unboldmath{} & \boldmath{}\textbf{0.14$\pm$0.01}\unboldmath{} \\
    \bottomrule
    \end{tabular}%
    \endgroup
	}
  \label{tab:results_stress_test}%
\end{table}%

\end{document}